\documentclass[journal]{IEEEtran}
\usepackage{amsmath,amsfonts}
\usepackage{algorithmic}
\usepackage{algorithm}
\usepackage{array}
\usepackage[caption=false,font=normalsize,labelfont=sf,textfont=sf]{subfig}
\usepackage{textcomp}
\usepackage{graphicx}
\usepackage{overpic} 
\usepackage{stfloats}
\usepackage{url}
\usepackage{verbatim}
\usepackage{tikz}
\usepackage{enumitem}
\usepackage{cite}
\usepackage{xcolor}

\usepackage{booktabs}
\usepackage{multirow}
\usepackage{makecell}
\usepackage[table]{xcolor}

\definecolor{bestcolor}{HTML}{D1D5F8}
\definecolor{secondcolor}{HTML}{E1F5FE}
\colorlet{mycitebg}{blue!10} 

\usepackage{hyperref}
\hypersetup{
    colorlinks=true,   
    linkcolor=blue,    
    citecolor=blue,    
    urlcolor=blue,     
    pdfborder={0 0 0}  
}

\begin{document}

\title{From Field to Planning: Learning Temporal Polar Gradient Fields for Vision-based AAV Trajectory Planning in Dynamic Environments}

\author{Haowen Gao$^{\dagger}$, Zhan Chen$^{\dagger}$, Yinuo Xu, Yifeng Niu, Mengyun Wang and Chao Yan 
\thanks{$^{\dagger}$These authors contributed equally to this work.

This work was supported in part by the National Natural Science Foundation of China under Grants 62576349, 62576354, 62533010, and in part by the Postgraduate Scientific Research Innovation Project of Hunan Province under Grant CX20240005. (\textit{Corresponding author: Yifeng Niu, Chao Yan.})

Haowen Gao, Zhan Chen, Yinuo Xu, Yifeng Niu and Mengyun Wang are with the College of Intelligence Science and Technology, National University of Defense Technology, Changsha 410073, China (e-mail: gaohaowen@nudt.edu.cn; chenzhan@nudt.edu.cn; xuyinuo@nudt.edu.cn; niuyifeng@nudt.edu.cn; wangmengyun@nudt.edu.cn).

Chao Yan is with the College of Automation Engineering, Nanjing University of Aeronautics and Astronautics, Nanjing 211106, China (e-mail: yanchao@nuaa.edu.cn).}}

\maketitle
\begin{abstract}
Autonomous Aerial Vehicle (AAV) trajectory planning requires real-time and accurate safety gradient guidance, especially in dynamic environments. Existing methods typically rely on static maps and dynamic obstacle tracking to construct gradient fields, imposing heavy computational burdens and yielding conservative trajectories.
To address these limitations, we propose a two-stage Field-to-Planning (F2P) learning framework to provide accurate and efficient safety gradients for trajectory planning in dynamic environments. In the first stage, we introduce an Ego-centric Polar Occupancy (EPO) representation that naturally matches the geometry of the onboard camera frustum, while supporting unstructured scene modeling. Subsequently, we formulate a Temporal Polar Gradient Field (TPGF) via multi-kernel separable Gaussian convolution to provide unified safety gradients, without explicitly distinguishing between static and dynamic obstacles. To mitigate the inherent spatial ambiguity introduced by applying convolutions within the polar grid, we propose an ESDF-guided offline learning strategy that incorporates geometric priors into TPGF. In the second stage, we employ the calibrated TPGF to guide the trajectory planning policy in a self-supervised learning manner. Comparative results with other approaches demonstrate the superiority of our method. Extensive real-world experiments within various cluttered dynamic environments validate the real-time performance and robustness on resource-constrained onboard systems.
\end{abstract}


\section{INTRODUCTION}
\IEEEPARstart{A}{utonomous} Aerial Vehicles (AAVs) have been increasingly employed in various applications, such as inspection, exploration, reconstruction, and autonomous navigation \cite{liu2024uav,zhang2023fgip,hua2025deep}. However, ensuring flight safety in unknown and cluttered environments remains a critical challenge, particularly in the presence of moving obstacles \cite{peng2026stage}. Unlike static or quasi-static scenes, dynamic environments are characterized by time-varying obstacle configurations that can rapidly invalidate previously planned trajectories. Consequently, AAVs must rapidly account for both static structures and moving obstacles from restricted local observations and perform frequent replanning, imposing substantial computational burdens on resource-constrained onboard systems.
\begin{figure}
    \centering
    \includegraphics[width=1.0\linewidth, keepaspectratio]{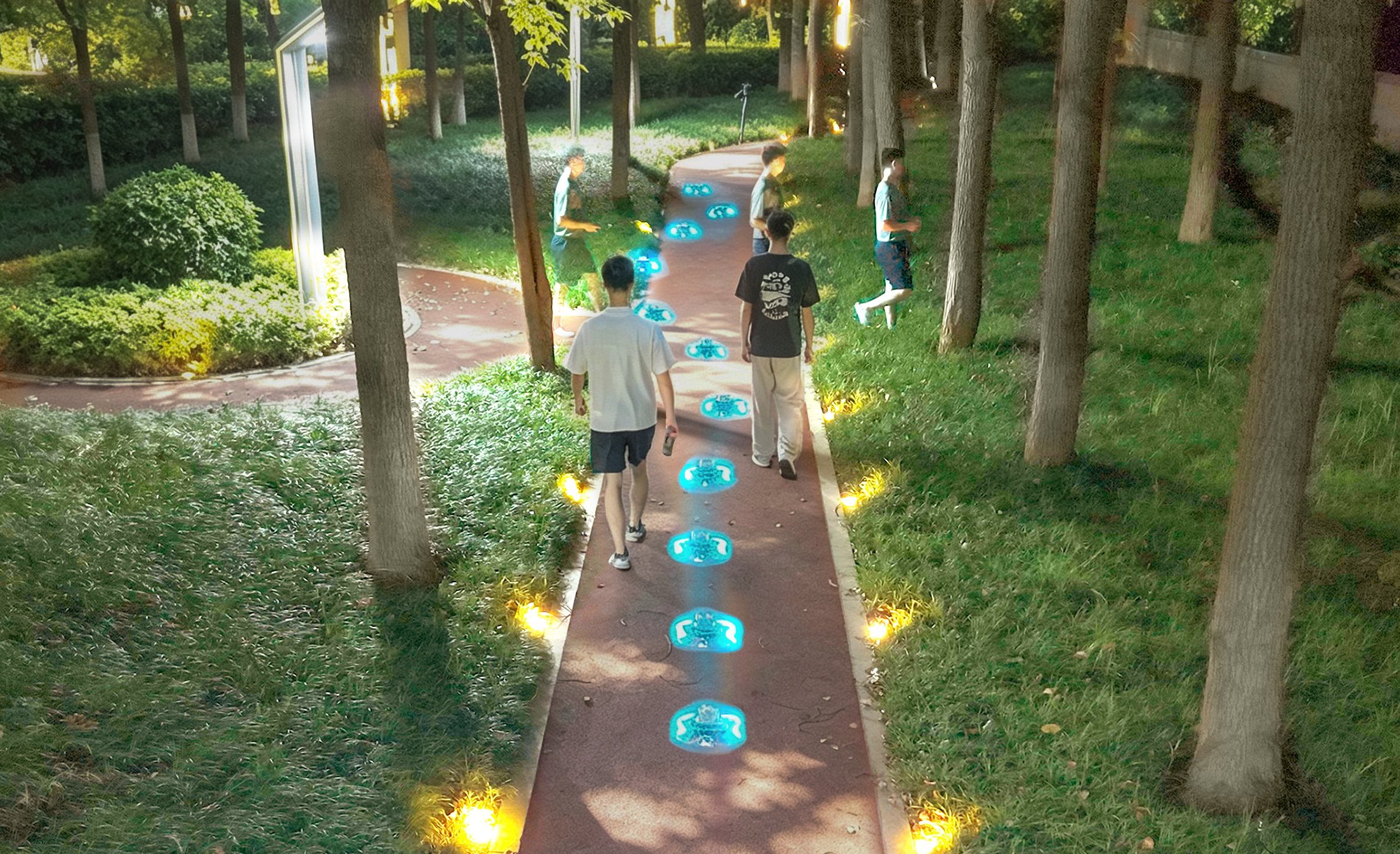}
    
    \vspace{-8pt} 
    \caption{Snapshot of the customized AAV and its trajectory flying through outdoor forest corridors using the proposed framework. The robot achieves safe autonomous flight, effectively executing collision-avoidance maneuvers against approaching pedestrians.}
    \label{fig:placeholder}
    \vspace{-12pt}
\end{figure}

Conventionally, modular pipelines\cite{chen2023risk,chen2023flying,xu2022vision} commonly leverage the detection and tracking of moving objects (DATMO) to track the dynamic obstacles and model them via bounding boxes or ellipsoids, applying distinct evasion strategies for static and dynamic threats. Although effective in avoiding simple dynamic obstacles, these methods might fail in highly cluttered environments\cite{xu2022vision}. Some works integrate explicit obstacle trajectory prediction into the planner \cite{zhu2019chance,lin2020robust}. However, they remain constrained by the unpredictability of highly dynamic obstacles and onboard sensing noise, suffering substantial performance degradation whenever the actual motion deviates from predefined models \cite{xu2025intent}. 

Learning-based approaches have gained attention in the field of obstacle avoidance due to the advantages of low latency and rapid response\cite{song2023learning,fan2025flying}. 
Several existing works\cite{fan2025flying},\cite{xu2025navrl} successfully employ LiDAR and deep reinforcement learning (DRL) for autonomous flight in dynamic environments. However, they suffer from extremely high training costs caused by low sample efficiency and time-consuming reward. 
Recent advancements\cite{yang2023iplanner,lu2023lpnet,lu2024you,han2025dynamically} focus on the self-supervised learning guided by privileged Euclidean Signed Distance Fields (ESDFs) for safety gradient evaluation. However, constructing and updating ESDFs for rapidly changing obstacle configurations is computationally expensive, which limits their applicability to self-supervised training in dynamic environments\cite{xu2022vision}. Moreover, utilizing global ESDFs leaks out-of-FoV privileged information and induces causal confusion. Thus, developing safety gradients aligned with partial visual inputs is required.

To address the above limitations, this paper proposes a two-stage Field-to-Planning (F2P) learning framework for vision-based AAV trajectory planning in dynamic environments. In the first stage, we propose an Ego-centric Polar Occupancy (EPO) representation, which naturally aligns with the perspective frustum of the depth camera. Constructed via the projection of privileged scene point clouds at future timestamps, EPO can represent various unstructured scenarios.
Leveraging the rectangular structural regularity of EPO, we generate a Temporal Polar Gradient Field (TPGF) via multi-kernel separable Gaussian convolution with range-bucketed angular compensation, without requiring explicit classification of static and dynamic obstacles. To mitigate the geometric ambiguity induced by applying convolutions within the polar grid, we introduce an ESDF-guided offline learning strategy, which effectively calibrates the gradient directions and endows the TPGF to learn explicit geometric priors. In the second stage, we employ the calibrated TPGF to guide the efficient self-supervised learning of the trajectory planning policy in dynamic environments.

The main contributions of this article are as follows:

\begin{enumerate}[label=\arabic*), leftmargin=*, nosep, after=\vspace{\baselineskip}]
\item We propose a two-stage vision-based learning framework for trajectory planning in dynamic environments, where a gradient field is first learned and subsequently utilized to guide the planning policy. Taking privileged scene point clouds as input, the proposed framework enables safety gradient guidance in various unstructured scenarios.
\item  We introduce a TPGF learning strategy, which constructs a differentiable safety gradient field via multi-scale Gaussian diffusion of ego-centric polar occupancies with a temporal channel to provide unified safety gradients across static and dynamic obstacles. Furthermore, we leverage an ESDF-guided offline learning to incorporate explicit geometric priors and improve gradient consistency. 
\item  We validate the proposed framework through extensive simulation and real-world experiments, demonstrating its superior performance compared to state-of-the-art (SOTA) approaches. Furthermore, it exhibits robust deployment capabilities in various cluttered dynamic environments.
\end{enumerate}
\section{RELATED WORK}
\subsection{Modular Approaches for Dynamic Obstacle Avoidance}
Obstacle avoidance in dynamic environments has long been a central focus of robotics research. Traditional solutions typically adopt a modular pipeline that decomposes the problem into perception, mapping, and planning\cite{lu2025yopov2}, leveraging the DATMO method to explicitly model static and dynamic obstacles, respectively\cite{xu2022vision,chen2021active}.
Map-based methods using environment representations such as ESDFs\cite{zhou2019robust} or occupancy voxels\cite{zhou2020ego} are widely adopted. Based on these representations, trajectory optimization techniques, including B-spline optimization\cite{zhou2019robust} and model predictive control (MPC)\cite{kamel2017robust},\cite{lin2020robust}, are employed to generate smooth and dynamically feasible trajectories. To handle moving obstacles, such planners incorporate a trajectory prediction module that models dynamic agents as spatiotemporal constraints. For example,\cite{xu2022vision},\cite{xu2025intent},\cite{lu2024fapp} explicitly estimate obstacle motion using dense point clouds. Despite improved fidelity, their high computational complexity restricts applicability in highly dynamic scenarios. Alternatively,\cite{chen2023flying} leverages RGB semantic priors to reduce this burden. However, this limits generalizability, and the auxiliary detection module still taxes constrained onboard resources.

While modular frameworks are reliable in sparse environments, their performance degrades significantly in densely cluttered scenarios due to perception latency and computational bottlenecks.

\subsection{Learning-Based Trajectory Planning}
Unlike traditional modular designs, end-to-end learning frameworks map raw sensor measurements directly into control signals, emerging as a highly effective paradigm for agile flight. For example,\cite{bhattacharya2025vision} utilizes Behavior Cloning to deduce a sensorimotor policy mapping depth inputs to velocity commands via expert supervision. While demonstrating impressive high-speed avoidance, this pure imitation restricts the system's upper bound to the expert's capabilities. To overcome these bottlenecks, recent works\cite{redder2025vds,kulkarni2024reinforcement,yu2024mavrl} apply DRL to iteratively refine flight policies via trial-and-error, acquiring robust navigation skills free from expert limitations. However, deploying DRL in aerial planning is notoriously plagued by low sample efficiency and challenging convergence due to continuous high-dimensional state spaces and sparse rewards. As an alternative,\cite{lu2023lpnet,lu2024you,lu2025yopov2} propose a differentiable self-supervised framework. By employing privileged ESDFs to explicitly optimize trajectory primitives via gradient descent, they ensure both interpretability and training efficiency in static scenes.

Nevertheless, ESDFs struggle to represent dynamic environments in real-time, and the high-dimensional output of multi-segment trajectories significantly exacerbates the convergence burden on the policy network. Distinct from these works, we propose a two-stage Field-to-Planning learning framework, utilizing a Temporal Polar Gradient Field to achieve robust and real-time trajectory planning in dynamic environments.
\begin{figure*}
    \centering
    \includegraphics[width=1.0\linewidth, keepaspectratio]{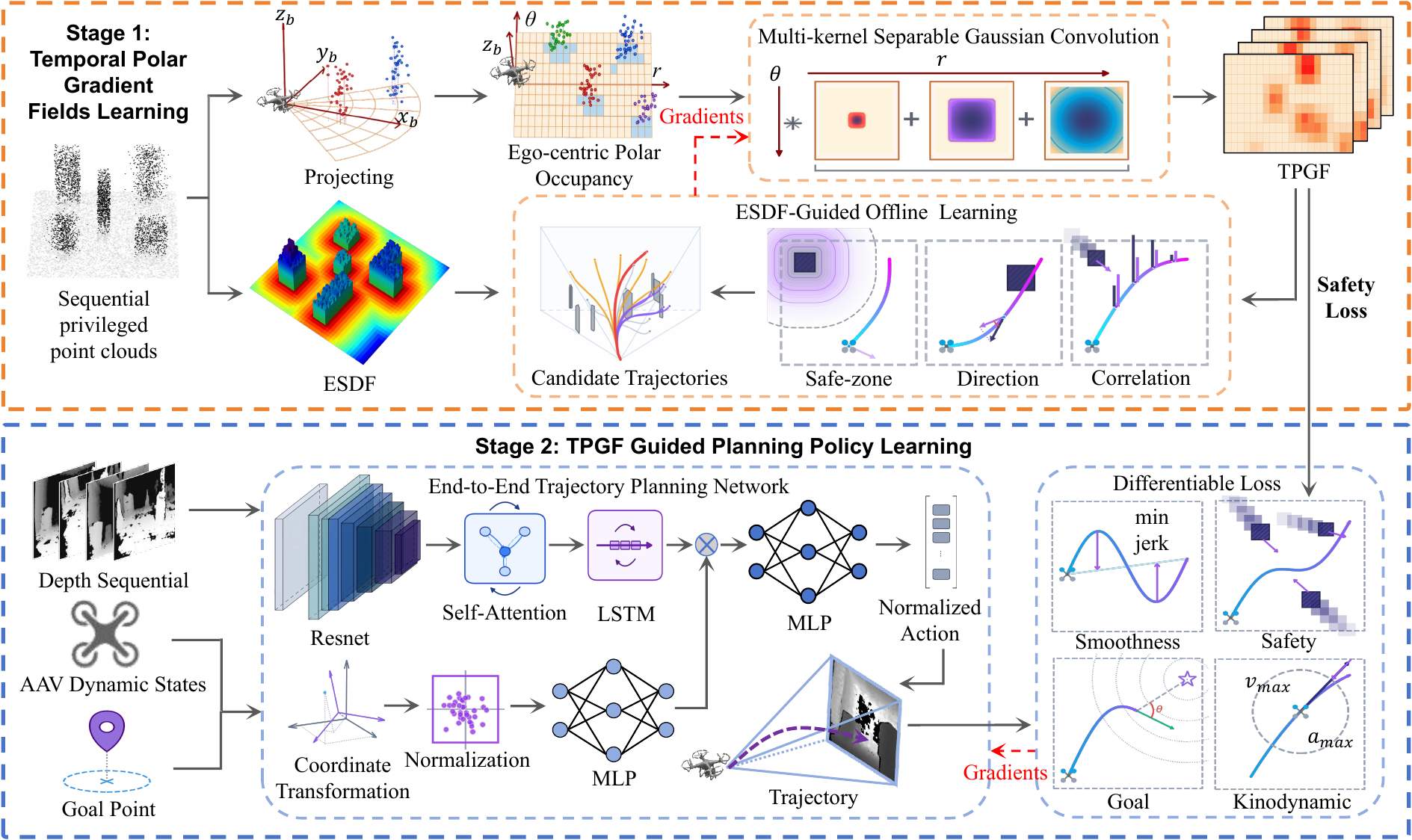}
    \vspace{-18pt}
    \caption{Overview of the proposed two-stage Field-to-Planning (F2P) framework. The first stage learns an ego-centric Temporal Polar Gradient Field (TPGF) from privileged point clouds. In the second stage, the calibrated TPGF is utilized as a differentiable safety field to guide the self-supervised learning of the trajectory planning policy.}
    \label{fig:network}
    \vspace{-8pt}
\end{figure*}
\section{PROBLEM FORMULATION}
This work addresses the problem of trajectory planning for AAVs within unknown dynamic environments. Relying solely on an RGB-D camera for obstacle perception, the AAV continuously plans smooth trajectories within its Field of View (FoV) toward the goal. The AAV operates using only onboard perception and state estimation. At any time $t$, the AAV has access to its estimated motion state and the goal, while the onboard camera provides a sequence of raw depth images with a limited sensing range. Let $\mathcal{I}$, $\mathcal{B}$, and $\mathcal{C}$ denote the world inertial frame, body frame, and camera frame, respectively. The camera is rigidly mounted on the airframe, and therefore the extrinsic transformation between $\mathcal{B}$ and $\mathcal{C}$ remains constant. 
The AAV state at the $k$-th planning step is denoted by
$\mathbf{x}_k = [\mathbf{p}_k, \mathbf{v}_k, \mathbf{a}_k, \mathbf{q}_k]^\top$,
where $\mathbf{p}_k \in \mathbb{R}^3$, $\mathbf{v}_k \in \mathbb{R}^3$, and $\mathbf{a}_k \in \mathbb{R}^3$ represent the position, linear velocity, and linear acceleration, respectively, while $\mathbf{q}_k \in \mathbb{S}^3$ denotes the attitude quaternion. The corresponding rotation matrix from the body frame to the world frame is denoted by $\mathbf{R}^{\mathcal{I}}_{\mathcal{B},k} \in SO(3)$. The goal at the same planning step is represented as $\mathbf{g}_k \in \mathbb{R}^3$ in the world frame. Let $\mathbf{d}_k$ denote the raw depth image captured by the onboard camera at the $k$-th sampling instant. The visual input is then defined as
$\mathcal{O}_k = \{\mathbf{d}_{k-n+1}, \dots, \mathbf{d}_{k-1}, \mathbf{d}_k\}$,
which contains $n$ depth frames sampled at a constant temporal interval $\Delta t$.

\section{METHODOLOGY}
We propose a two-stage F2P learning framework for vision-based AAV trajectory planning in dynamic environments. The overview of the proposed framework is illustrated in Fig. \ref{fig:network}. Our F2P learning framework mainly consists of two stages: temporal polar gradient fields learning stage and TPGF guided planning policy learning stage. In stage 1, we construct ego-centric TPGFs from sequential privileged point clouds and introduce an ESDF-guided offline learning strategy to distill explicit geometric priors into the field. In stage 2, the learned TPGF is used as a differentiable safety field to guide the self-supervised learning of the planning policy in dynamic environments.

\subsection{Temporal Polar Gradient Field Learning}
\begin{figure}
    \centering
    \includegraphics[width=1.0\linewidth, keepaspectratio]{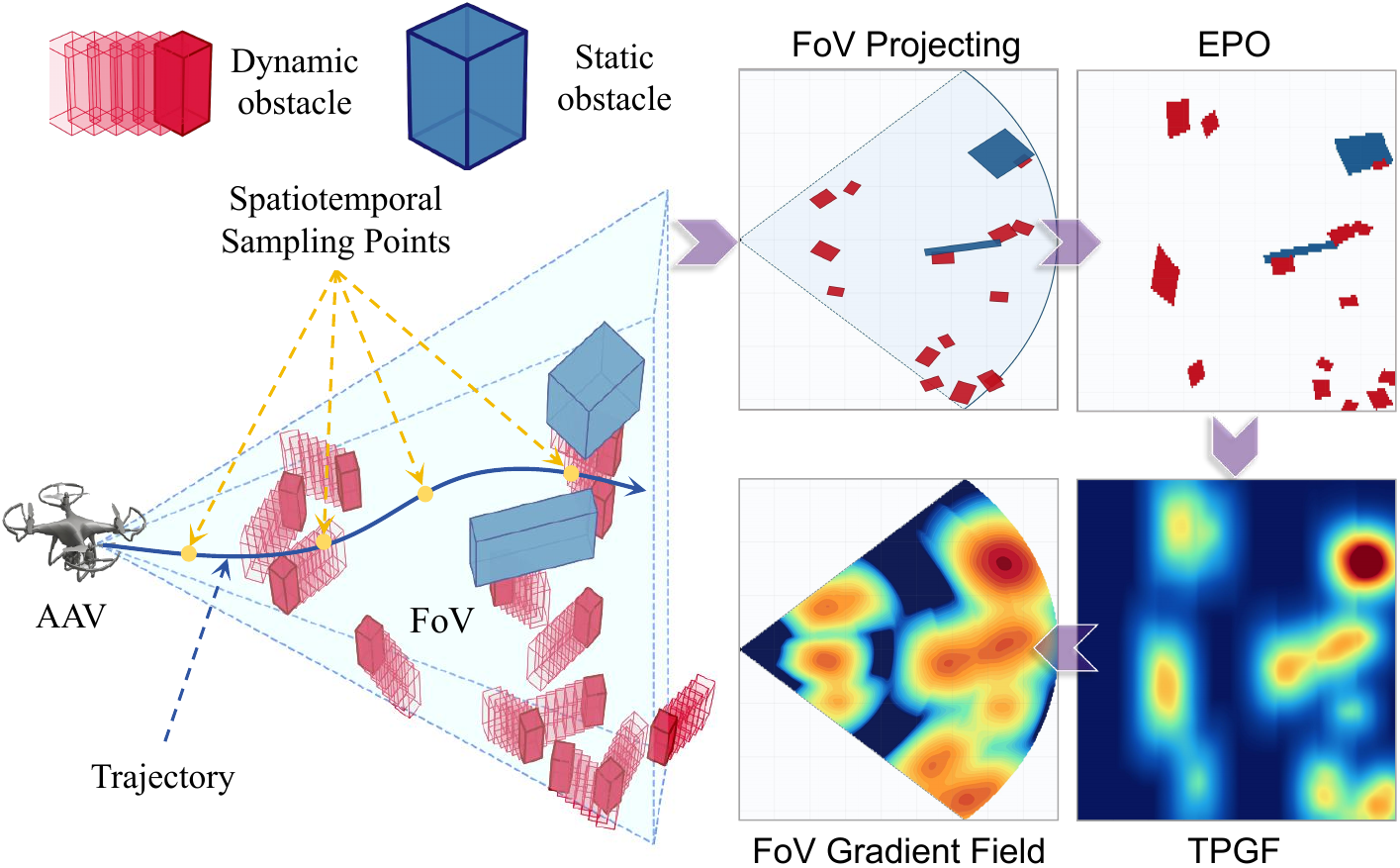}
    \vspace{-15pt}
    \caption{Illustration of the TPGF generation and trajectory evaluation process. Blue and red cubes represent static and moving obstacles, respectively. First, obstacle points within the FoV are projected into the polar coordinate grid to construct the EPO, followed by a convolution operation to generate the TPGF. Finally, spatiotemporal sampling points along the planned trajectory are projected into the corresponding TPGF slices to query risk values, thereby yielding the differentiable safety loss.}
    \label{fig:scene}
    \vspace{-10pt}
\end{figure}

In this section, we formulate the generation and ESDF-guided learning of the TPGF. As shown in Fig. \ref{fig:scene}, privileged in-FoV point clouds are first projected to construct the EPO, which is subsequently transformed into a differentiable TPGF via multi-kernel separable Gaussian convolution with range-bucketed angular compensation. Finally, an ESDF-guided offline learning strategy is applied to align the TPGF with explicit geometric priors.
\subsubsection{Ego-centric Polar Occupancy Representation}
We first use privileged world-frame point clouds sampled from obstacles to construct an EPO representation. For the $\tau$-th spatiotemporal sampling point, let
$\mathcal{P}^{\mathcal{I}}_{k,\tau}=\{\mathbf{s}^{\mathcal{I}}_{k,\tau,l}\in\mathbb{R}^3\}_{l=1}^{N_{k,\tau}}$ denote the privileged point cloud in the inertial frame. Given the AAV position $\mathbf{p}^{\mathcal{I}}_k$ and rotation matrix
$\mathbf{R}^{\mathcal{I}}_{\mathcal{B},k}$, each privileged point is first transformed into the body frame:
\begin{equation}
\mathbf{s}^{\mathcal{B}}_{k,\tau,l}
=
\left(\mathbf{R}^{\mathcal{I}}_{\mathcal{B},k}\right)^\top
\left(
\mathbf{s}^{\mathcal{I}}_{k,\tau,l}
-
\mathbf{p}^{\mathcal{I}}_k
\right).
\label{body}
\end{equation}
where $\mathbf{s}^{\mathcal{B}}_{k,\tau,l}
=
[x^{\mathcal{B}}_{k,\tau,l},
y^{\mathcal{B}}_{k,\tau,l},
z^{\mathcal{B}}_{k,\tau,l}]^\top$.
Only the points inside the onboard FoV are retained:
\begin{align}
\mathcal{P}^{\mathcal{B},F}_{k,\tau}
=
\Bigg\{
\mathbf{s}^{\mathcal{B}}_{k,\tau,l}
\;\Bigg|\;
& 0 < x^{\mathcal{B}}_{k,\tau,l} \le r_{\max},
\;
z_{\min} \le z^{\mathcal{B}}_{k,\tau,l} \le z_{\max}, \nonumber \\
& -\frac{\phi}{2} \le \tan^{-1} \frac{y^{\mathcal{B}}_{k,\tau,l}}{x^{\mathcal{B}}_{k,\tau,l}} < \frac{\phi}{2}
\Bigg\},
\end{align}
where $r_{\max}$ is the sensing range, $\phi$ is the horizontal FoV angle, and $[z_{\min}, z_{\max}]$ defines the vertical filtering boundaries.

We adopt a FoV-aligned ego-centric polar occupancy representation. Specifically, the retained in-FoV points are projected onto the local $xy$-plane and mapped into the 2D polar domain:
\begin{equation}
\begin{aligned}
r_{k,\tau,l} 
&= 
\sqrt{
\left(x^{\mathcal{B}}_{k,\tau,l}\right)^2
+
\left(y^{\mathcal{B}}_{k,\tau,l}\right)^2
}, 
\\
\theta_{k,\tau,l} 
&= 
\operatorname{atan2}
\left(
y^{\mathcal{B}}_{k,\tau,l}, \; x^{\mathcal{B}}_{k,\tau,l}
\right).
\end{aligned}
\end{equation}

The coordinate transformation is smooth, invertible, and fully differentiable. To process the unordered point cloud efficiently via neural networks, the continuous polar coordinates are discretized into a fixed-size 2D binary tensor. Given $N_r$ radial bins and $N_\theta$ angular bins, the corresponding spatial resolutions are defined as $\Delta r = r_{\max}/N_r$ and $\Delta \theta = \phi/N_\theta$. The EPO map $\mathbf{E}_{k,\tau} \in \{0,1\}^{N_r \times N_\theta}$ is then constructed using the following indicator function:
\begin{equation}
\mathbf{E}_{k,\tau}(u,v)
=
\mathbb{I}
\left[
\exists l,\;
u=\left\lfloor \frac{r_{k,\tau,l}}{\Delta r} \right\rfloor,
\;
v=\left\lfloor \frac{\theta_{k,\tau,l}+\phi/2}{\Delta \theta} \right\rfloor
\right],
\label{eq:EPO}
\end{equation}
where $\mathbb{I}[\cdot]$ evaluates to $1$ if at least one obstacle point falls within the spatial span of the $(u,v)$-th cell, and $0$ otherwise, and $\lfloor \cdot \rfloor$ denotes the floor function.

\subsubsection{Temporal Polar Gradient Field Construction}
Using multi-kernel separable Gaussian convolution with range-bucketed angular compensation, we transform the binary polar occupancy map into a smooth, differentiable TPGF.

Specifically, the FoV-aligned EPO representation is processed through three
parallel Gaussian convolution layers
$\ell \in \{\mathrm{core},\mathrm{body},\mathrm{halo}\}$,
parameterized by progressively increasing learnable base bandwidths.
Each layer $\ell$ contains a single learnable Gaussian base bandwidth
$s_\ell$, measured in polar-grid cells, which is shared by the radial
kernel and the range-compensated angular kernel.

For each layer $\ell$, we first convolve the FoV-aligned EPO map along
the radial dimension:
\begin{equation}
\mathbf{H}_{\ell,k,\tau}
=
\mathbf{G}^{r}_{\ell} *_{r} \mathbf{E}_{k,\tau},
\label{radial}
\end{equation}
where $*_{r}$ denotes 1D convolution along the radial axis. The discrete
radial Gaussian kernel is defined as
\begin{equation}
\mathbf{G}^{r}_{\ell}(i)
=
\frac{1}{Z_{\ell}^{r}}
\exp\left(
-\frac{i^2}{2s_{\ell}^{2}}
\right),
\qquad
i\in[-K_{\ell}^{r},K_{\ell}^{r}],
\end{equation}
where $Z_{\ell}^{r}$ is the normalization factor and
$K_{\ell}^{r}=\lceil 3s_{\ell}\rceil$ is the truncated radial kernel
radius based on the three-sigma rule.

The radial axis is divided into $M$ buckets
$\{\mathcal{R}_m\}_{m=1}^{M}$. To compensate for the range-dependent
distortion of the polar grid, we define the angular scaling factor for
the $m$-th radial bucket as
\begin{equation}
c_m =
\operatorname{clip}\left(
\frac{r_{\mathrm{ref}}}
{\max(\bar r_m,r_{\min})},
c_{\min},c_{\max}
\right),
\qquad
s_{\ell,m}^{\theta}=s_\ell c_m ,
\end{equation}
where $\bar r_m$ is the center distance of the $m$-th bucket,
$r_{\mathrm{ref}}$ is the reference distance, and
$r_{\min}$, $c_{\min}$, and $c_{\max}$ are fixed hyperparameters.
Therefore, $s_{\ell,m}^{\theta}$ is deterministically derived from
$s_\ell$ rather than independently learned.

The corresponding discrete angular Gaussian kernel is formulated as
\begin{equation}
\mathbf{G}^{\theta}_{\ell,m}(j)
=
\frac{1}{Z_{\ell,m}^{\theta}}
\exp\left(
-\frac{j^2}{2(s_{\ell,m}^{\theta})^2}
\right),
\qquad
j\in[-K_{\ell}^{\theta},K_{\ell}^{\theta}],
\end{equation}
where $Z_{\ell,m}^{\theta}$ is the normalization factor and
$K_{\ell}^{\theta}$ denotes the truncated angular kernel radius for
layer $\ell$.

The corresponding standard deviations in physical radial and angular
coordinates are
$\sigma_{\ell}^{r}=s_{\ell}\Delta r$ and
$\sigma_{\ell,m}^{\theta}=s_{\ell,m}^{\theta}\Delta\theta$,
respectively.

\begin{equation}
\mathbf{S}_{k,\tau} = \eta \sum_{\ell} w_\ell \sum_{m=1}^{M} \mathbf{M}_m \odot \left( \mathbf{G}^{\theta}_{\ell,m} *_{\theta} \mathbf{H}_{\ell,k,\tau} \right),
\label{eq:PGF}
\end{equation}
where $\eta$ is the threat scaling factor, $*_{\theta}$ denotes 1D convolution along the angular axis, $\mathbf{M}_m$ is the binary mask selecting polar cells within $\mathcal{R}_m$ from the FoV-aligned EPO, and $\odot$ denotes element-wise multiplication.

\subsubsection{ESDF-Guided Offline Learning}
While the proposed TPGF inherently generates repulsive gradients pointing away from obstacles, it lacks explicit metric scaling, leading to inherent spatial ambiguity. In contrast, ESDFs formulate gradients based on exact metric distances, providing precise and intuitive geometric priors for collision avoidance. Thus, we introduce an ESDF-guided offline learning strategy. Rather than training the planning policy, this module distills geometric
priors from ESDFs to optimize a compact set of TPGF structural parameters
$\Theta =
\{\eta,\{s_\ell,w_\ell\}_{\ell\in
\{\mathrm{core},\mathrm{body},\mathrm{halo}\}}\}$,
where $s_\ell$ denotes the shared learnable base bandwidth of layer
$\ell$ in polar-grid cells, and $w_\ell$ denotes the corresponding kernel
weight. The bucket-specific angular bandwidth $s_{\ell,m}^{\theta}$ is
deterministically derived from $s_\ell$ and is not independently optimized.

To prevent distribution shift and emphasize high-risk regions, we sample a diverse candidate trajectory set $\mathcal{T}_k$ encompassing policy, anchor, random, and near-obstacle trajectories
. This is formulated as an optimization problem to find the optimal structural parameters $\Theta^*$:
\begin{equation}
\begin{split}
\Theta^* = \arg\min_{\Theta} \Big( &\lambda_{\mathrm{corr}}\mathcal{L}_{\mathrm{corr}} + \lambda_{\mathrm{safe}}\mathcal{L}_{\mathrm{safe}} \\
&+ \lambda_{\mathrm{grad}}\mathcal{L}_{\mathrm{grad}} + \lambda_{\mathrm{reg}}\mathcal{L}_{\mathrm{reg}} \Big).
\end{split}
\label{eq:loss_calib}
\end{equation}
where $\mathcal{L}_{\mathrm{corr}}$ is the correlation loss, $\mathcal{L}_{\mathrm{safe}}$ is the safe-zone loss, $\mathcal{L}_{\mathrm{grad}}$ is the gradient loss, and $\mathcal{L}_{\mathrm{reg}}$ is the regularization loss. We solve this objective using gradient-based optimization, iteratively updating the multi-scale Gaussian kernel parameters.

Specifically, the $\mathcal{L}_{\mathrm{corr}}$ preserves 
the relative risk ranking without enforcing strict numerical equality:
\begin{equation}
\mathcal{L}_{\mathrm{corr}} = 1 - \mathrm{Corr}_{\xi \in \mathcal{T}_k} \big( C_\Theta(\xi), C_T(\xi) \big),
\label{eq:loss_corr}
\end{equation}
where the student TPGF cost $C_\Theta(\xi)$ and the teacher ESDF cost $C_T(\xi)$ 
are computed via discrete path integration over the waypoints:
\begin{equation}
C_\Theta(\xi) = \sum_{\tau} \mathbf{S}^2(\mathbf{p}_\tau) \Delta t, \quad
C_T(\xi) = \sum_{\tau} \mathcal{C}_{\mathrm{ESDF}}(\mathbf{p}_\tau) \Delta t.
\end{equation}
where $\mathbf{S}(\mathbf{p}_\tau)$ and $\mathcal{C}_{\mathrm{ESDF}}(\mathbf{p}_\tau)$ 
represent the TPGF risk value and the ESDF safety penalty 
queried at the trajectory waypoint $\mathbf{p}_\tau$, respectively.

Furthermore, to prevent overly conservative trajectory planning, 
the safe-zone loss $\mathcal{L}_{\mathrm{safe}}$ suppresses false-positive penalties 
in teacher-safe regions:
\begin{equation}
\mathcal{L}_{\mathrm{safe}} = \mathbb{E}_{\xi \in \mathcal{T}_k} \Big[ \omega_T(\xi) \left[ C_\Theta(\xi) - \tau_P \right]_+^2 \Big],
\label{eq:loss_safe}
\end{equation}
where $\omega_T(\xi)$ is a soft weighting function that activates 
exclusively when the teacher cost is below a safety threshold.

Crucially, to ensure the TPGF provides reliable repulsive directions 
for the planner, the gradient loss $\mathcal{L}_{\mathrm{grad}}$ enforces both 
directional consistency and magnitude alignment between the student 
and teacher safety gradients (denoted as $\mathbf{g}_\Theta$ and $\mathbf{g}_T$, respectively):
\begin{equation}
\begin{split}
\mathcal{L}_{\mathrm{grad}} &= \mathbb{E}_{\xi \in \mathcal{T}_k} \Big[ \big[ m - \cos(\mathbf{g}_\Theta, \mathbf{g}_T) \big]_+^2 \\
&\quad + \mathcal{H} \big( \log(1+\|\mathbf{g}_\Theta\|), \log(1+\|\mathbf{g}_T\|) \big) \Big],
\end{split}
\label{eq:loss_grad}
\end{equation}
where $m$ is the desired cosine margin and $\mathcal{H}$ denotes the Huber loss. 
Finally, a logarithmic regularization term $\mathcal{L}_{\mathrm{reg}}$ is applied 
to prevent the calibrated parameters from drifting excessively 
from their empirical initializations.

\subsection{TPGF-Guided Planning Policy Learning}
Guided by the calibrated gradient field, we optimize the planning policy in a self-supervised manner. This approach leverages privileged scene information, offering the advantage of a fully differentiable mapping from the network outputs to the loss function\cite{yang2023iplanner},\cite{lu2024you}. This section first introduces our trajectory planning network, and then details the carefully designed differentiable loss components, including the core TPGF-guided safety loss.
\subsubsection{End-to-End Trajectory Planning Network}
Our planning network extracts and fuses visual features and kinematic states to plan spatiotemporal polynomial trajectories. For the visual stream, a sequence of depth images $\mathcal{O}_k = \{\mathbf{d}_{k-n+1}, \dots, \mathbf{d}_k\}$ is processed to implicitly capture the motion patterns of dynamic obstacles. We employ a shared ResNet backbone for spatial feature extraction, followed by a Multi-Head 
Self-Attention (MHSA) module and an long short-term memory (LSTM) network. The entire visual encoding process is concisely formulated as:
\begin{equation}
\mathbf{h}_{\mathrm{vis}} = \mathrm{LSTM} \big( \mathrm{MHSA}(\mathrm{ResNet}(\mathcal{O}_k)) \big).
\label{lstm}
\end{equation}

Concurrently, the kinematic stream processes the AAV dynamic states and 
the goal point. To ensure spatial consistency, these raw inputs are 
transformed into the local body frame $\mathcal{B}$ and normalized, 
yielding the state vector $\tilde{\mathbf{s}}_k$. A Multi-Layer Perceptron (MLP) 
then encodes this into a compact kinematic representation $\mathbf{h}_{\mathrm{kin}}$:
\begin{equation}
\mathbf{h}_{\mathrm{kin}} = \mathrm{MLP}_{\mathrm{kin}}(\tilde{\mathbf{s}}_k).
\end{equation}

The visual and kinematic representations are concatenated and fed into 
a policy network to predict the normalized action $\mathbf{u}_k$:
\begin{equation}
\mathbf{u}_k = \mathrm{MLP}_{\mathrm{policy}} \big( [\mathbf{h}_{\mathrm{vis}} \parallel \mathbf{h}_{\mathrm{kin}}] \big),
\label{mlp}
\end{equation}
\begin{algorithm}
\caption{F2P Learning Procedure}
\label{alg:f2p_training}
\textbf{Input:} \parbox[t]{\dimexpr\linewidth-3.5em}{%
    Training set $\mathcal{D}_{\mathrm{train}}$, validation set $\mathcal{D}_{\mathrm{val}}$,
    max epochs $E$, batch size $B$, planning network $\pi_{\boldsymbol{\psi}}$,
    calibrated TPGF parameters $\Theta$, and spatiotemporal sampling points $N_\tau$.
}
\begin{algorithmic}[1]
\STATE Initialize the network parameters $\boldsymbol{\psi}$ and learning rate $\alpha$
\FOR{$e = 1$ to $E$}
    \FOR{each batch of $B$ samples from $\mathcal{D}_{\mathrm{train}}$}
        \STATE Extract AAV states $\tilde{\mathbf{s}}$, image sequences $\mathcal{O}$, and privileged obstacle points $\mathcal{P}^{\mathcal{I}}$
        \STATE Construct batched EPO $\mathbf{E}$ with Eq.~\eqref{body}-\eqref{eq:EPO}
        \STATE Generate calibrated TPGF $\mathbf{S}$ with Eq.~\eqref{radial}-\eqref{eq:PGF}
        \STATE Inference $\pi_{\boldsymbol{\psi}}(\mathcal{O}, \tilde{\mathbf{s}})$ with Eq.~\eqref{lstm}-\eqref{mlp}
        \STATE Sample trajectory states $\smash{\{\mathbf{p}_n,\mathbf{v}_n,\mathbf{a}_n,\mathbf{j}_n\}_{n=1}^{N_\tau}}$
        \STATE Query risk values $\smash{\boldsymbol{\rho}}$ along sampled trajectories
        \STATE Compute $\mathcal{L}_{\mathrm{safety}}$ with Eq.~\eqref{PGF_loss}-\eqref{eq:safety_loss}
        \STATE Compute $\mathcal{L}_{\mathrm{goal}}$, $\mathcal{L}_{\mathrm{smooth}}$ and $\mathcal{L}_{\mathrm{kino}}$ with Eq.~\eqref{eq:goal_loss}, \eqref{eq:smooth_loss}, and \eqref{eq:kino_loss}
        \STATE Update the network parameters by $\boldsymbol{\psi} \leftarrow \boldsymbol{\psi} - \alpha \nabla_{\boldsymbol{\psi}} \mathcal{L}$
    \ENDFOR
    \STATE Evaluate metrics on $\mathcal{D}_{val}$, step learning rate scheduler
\ENDFOR
\end{algorithmic}
\end{algorithm}
where $\parallel$ denotes the concatenation operation. Finally, the normalized action $\mathbf{u}_k$ is deterministically decoded into a fifth-order spatiotemporal polynomial trajectory $\xi_k$ with real physical scales to guide the autonomous flight\cite{lu2023lpnet},\cite{lu2024you}.

\subsubsection{Differentiable Loss}
Our carefully designed differentiable loss function is formulated as a weighted combination of four distinct components: goal guidance, collision safety, trajectory smoothness, and kinodynamic feasibility:
\begin{equation}
\min_{\boldsymbol{\psi}} \mathcal{L}(\boldsymbol{\psi}) = \lambda_{g} \mathcal{L}_{\mathrm{goal}} + \lambda_{a} \mathcal{L}_{\mathrm{safety}} + \lambda_{s} \mathcal{L}_{\mathrm{smooth}} + \lambda_{k} \mathcal{L}_{\mathrm{kino}},
\label{eq:total_loss}
\end{equation}
where $\boldsymbol{\psi}$ denotes the weights of the planning policy network, and $\lambda_{g}$, $\lambda_{a}$, $\lambda_{s}$, $\lambda_{k}$ are the predefined hyper-parameters balancing the different objectives.

Specifically, to ensure a uniform spatial scale for the objective 
distance throughout the learning process, the local goal $\mathbf{g}_{\mathrm{local}}$ 
is first determined by projecting a fixed guidance distance $D_{\mathrm{guide}}$ 
from the AAV's current position $\mathbf{p}_0$ along the direction 
toward the global target $\mathbf{g}_k$. The positional term $\mathcal{L}_{\mathrm{pos}}$ minimizes the spatial deviation of the terminal 
state $\mathbf{p}_f$ from the generated local goal $\mathbf{g}_{\mathrm{local}}$, 
while the velocity term $\mathcal{L}_{\mathrm{vel}}$ enforces directional 
consistency by aligning the normalized terminal velocity $\mathbf{v}_f$ with 
the target vector pointing from $\mathbf{p}_f$ to $\mathbf{g}_{\mathrm{local}}$. 
Formally, this unified objective is formulated as:
\begin{equation}
\begin{aligned}
    \mathbf{g}_{\mathrm{local}} &= \mathbf{p}_0 + D_{\mathrm{guide}} \frac{\mathbf{g}_k - \mathbf{p}_0}{\| \mathbf{g}_k - \mathbf{p}_0 \|}, \\
    \mathcal{L}_{\mathrm{pos}} &= \mathcal{L}_{S1}\left( \mathbf{p}_f, \mathbf{g}_{\mathrm{local}} \right), \\
    \mathcal{L}_{\mathrm{vel}} &= \mathcal{L}_{S1}\left( \frac{\mathbf{v}_f}{\|\mathbf{v}_f\| + \epsilon}, \frac{\mathbf{g}_{\mathrm{local}} - \mathbf{p}_f}{\| \mathbf{g}_{\mathrm{local}} - \mathbf{p}_f \| + \epsilon} \right), \\
    \mathcal{L}_{\mathrm{goal}} &= \lambda_{\mathrm{pos}} \mathcal{L}_{\mathrm{pos}} + \lambda_{\mathrm{align}} \mathcal{L}_{\mathrm{vel}},
\end{aligned}
\label{eq:goal_loss}
\end{equation}
where $\mathcal{L}_{S1}(\cdot)$ denotes the Smooth L1 loss function, 
$\lambda_{\mathrm{pos}}$ and $\lambda_{\mathrm{align}}$ are predefined hyper-parameters, and $\epsilon$ is a small constant ensuring numerical stability.

The calibrated TPGF $\mathbf{S}_{k,\tau}$ in Eq.~\eqref{eq:PGF} essentially serves as a differentiable risk map parameterized in polar coordinates $(r, \theta)$. For the $k$-th trajectory generated by the 
policy, we sample $N_\tau$ spatiotemporal points along the curve. These points are then projected 
into the polar domain to query the corresponding risk values, which 
are subsequently integrated over time to obtain the safety loss:
\begin{equation}
    \mathcal{L}_{\mathrm{TPGF}} = \sum_{\tau=1}^{N_\tau} \gamma^{\tau-1} \rho_{k,\tau}^2 \delta t,
\label{PGF_loss}
\end{equation}
where $\gamma \in (0, 1]$ is the temporal discount factor, and 
$\rho_{k,\tau} = \operatorname{Interp}\left(\mathbf{S}_{k,\tau}, r_{k,\tau}, \theta_{k,\tau}\right)$ 
denotes the continuous risk value obtained via differentiable bilinear 
interpolation on the corresponding $\mathbf{S}_{k,\tau}$ using the  polar coordinates $(r_{k,\tau}, \theta_{k,\tau})$ of the $\tau$-th trajectory point. 
\begin{figure}
    \centering
    \begin{minipage}{0.4939\linewidth}
        \centering
        \includegraphics[width=\linewidth, height=3.2cm]{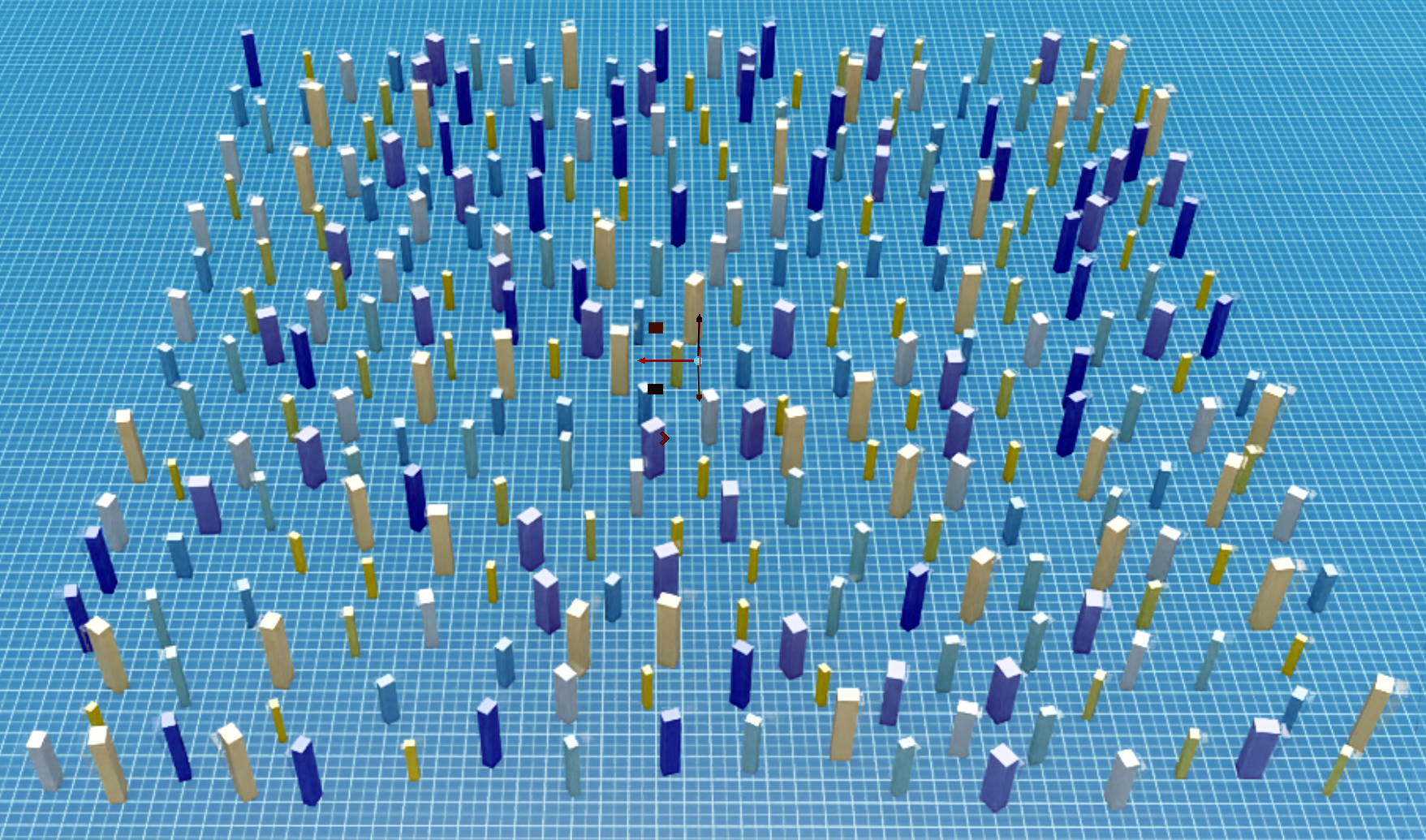}
        \vspace{0pt} 
        \centerline{\scriptsize (a) Isaac Sim Training Scenarios}
    \end{minipage}
    \hfill 
    \begin{minipage}{0.4939\linewidth}
        \centering
        \includegraphics[width=\linewidth, height=3.2cm]{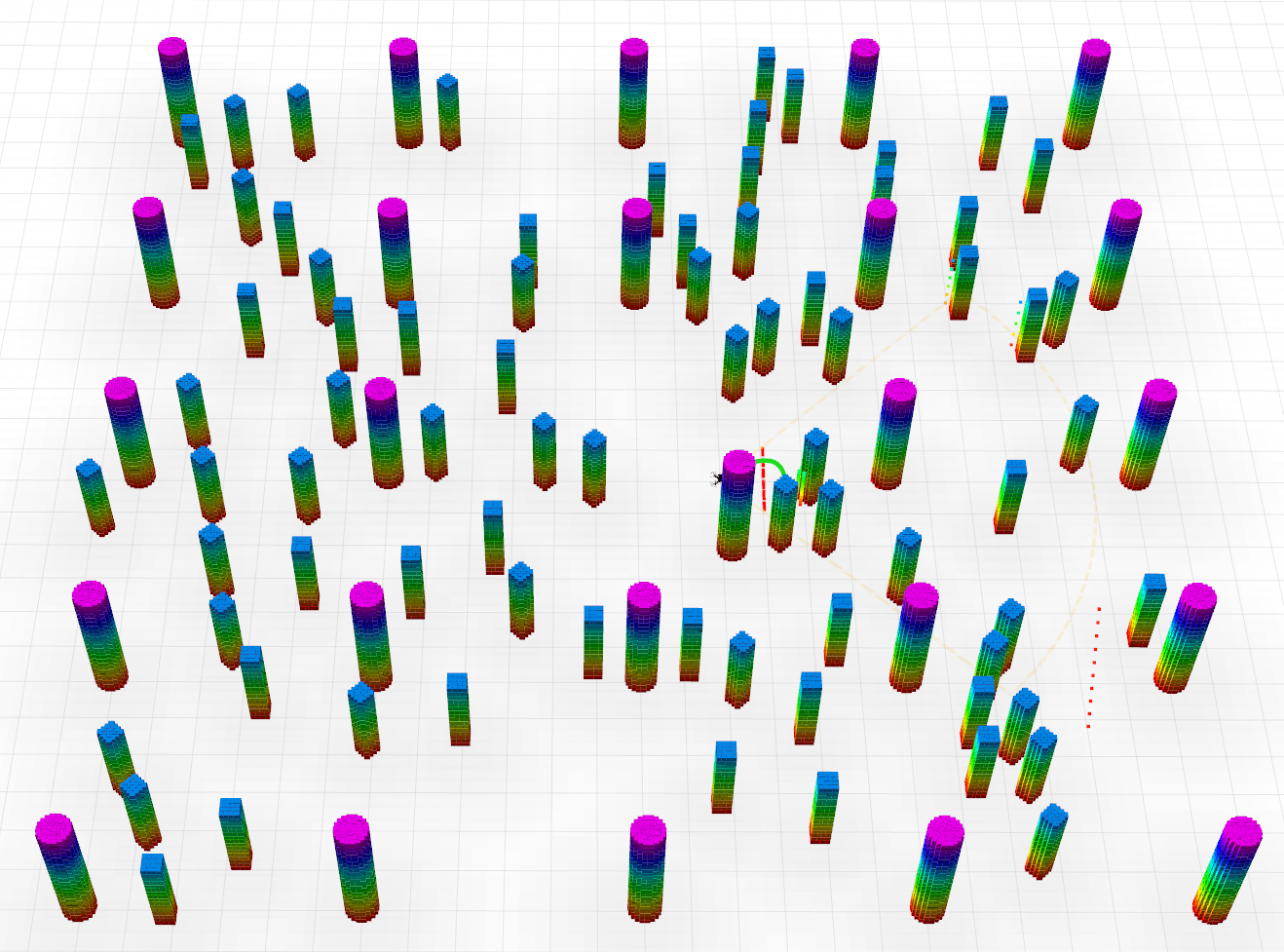}
        \vspace{0pt}
        \centerline{\scriptsize (b) Gazebo Test Scenarios}
    \end{minipage}
    
    \caption{Offline data collection scenarios for training and testing scenarios for planning benchmark comparisons.}
    \label{fig:scenarios}
    \vspace{-10pt}
\end{figure}

\begin{table}
    \centering
    \caption{Computational Efficiency Comparison between TPGF and ESDF}
    \label{tab:computation_efficiency}

    \renewcommand{\arraystretch}{1.25} 
    \setlength{\tabcolsep}{5.0pt}      
    
    \begin{tabular}{lcccc}
    \toprule
    
    \begin{tabular}{@{}c@{}}\textbf{Config.} \\ ($B \times N_\tau$)\end{tabular} & 
    \textbf{Frames} & 
    \begin{tabular}{@{}c@{}}\textbf{TPGF} \\  \textbf{Runtime} \\ (ms)\end{tabular} & 
    \begin{tabular}{@{}c@{}}\textbf{ESDF} \\ \textbf{Runtime} \\ (ms)\end{tabular} & 
    \begin{tabular}{@{}c@{}}\textbf{Speedup} \\ (ESDF / TPGF)\end{tabular} \\
    
    \midrule
    
    $1 \times 1$    & 1    & 0.454  & 1.854   & 4.09$\times$   \\
    $8 \times 30$   & 240  & 1.093  & 351.69  & 321.91$\times$ \\
    $32 \times 30$  & 960  & 3.515  & 1438.02 & 409.36$\times$ \\
    $128 \times 30$ & 3840 & 14.076 & 5695.21 & 404.90$\times$ \\
    
    \bottomrule
    \end{tabular}
    
    \vspace{2pt}
    \vspace{-15pt}
\end{table}

Since all operations are fully differentiable, the spatial safety gradient can be backpropagated to the policy parameters $\boldsymbol{\psi}$ via the chain rule:\par
\begin{footnotesize}
\begin{equation}
    \frac{\partial \mathcal{L}_{\mathrm{TPGF}}}{\partial \boldsymbol{\psi}} = \sum_{\tau=1}^{N_\tau} \frac{\partial \mathcal{L}_{\mathrm{TPGF}}}{\partial \rho_{k,\tau}} \left( \frac{\partial \operatorname{Interp}}{\partial r_{k,\tau}} \frac{\partial r_{k,\tau}}{\partial \mathbf{p}_{k,\tau}} + \frac{\partial \operatorname{Interp}}{\partial \theta_{k,\tau}} \frac{\partial \theta_{k,\tau}}{\partial \mathbf{p}_{k,\tau}} \right) \frac{\partial \mathbf{p}_{k,\tau}}{\partial \boldsymbol{\psi}},
    \label{eq:safety_grad}
\end{equation}
\end{footnotesize}

To comprehensively address sensor limitations and spatial bounds, 
this TPGF loss is further combined with FoV and height constraints to form the total safety objective:
\begin{equation}
    \mathcal{L}_{\mathrm{safety}} = \lambda_p \mathcal{L}_{\mathrm{TPGF}} + \lambda_f \mathcal{L}_{\mathrm{FoV}} + \lambda_z \mathcal{L}_{\mathrm{height}}.
\label{eq:safety_loss}
\end{equation}
where $\lambda_p$, $\lambda_f$, and $\lambda_z$ are the weighting 
coefficients. The FoV loss $\mathcal{L}_{\mathrm{FoV}}$ enforces the planned trajectory to remain within the valid observation range, while the height loss $\mathcal{L}_{\mathrm{height}}$ restricts the trajectory within a reasonable altitude range.

To ensure stable onboard perception and facilitate reliable obstacle observation, the smoothness loss $\mathcal{L}_{\mathrm{smooth}}$ is formulated to minimize jerk: 
\begin{equation}
    \mathcal{L}_{\mathrm{smooth}} = \frac{1}{N_\tau} \sum_{\tau=1}^{N_\tau} \| \dddot{\mathbf{p}}_{\tau} \|^2,
    \label{eq:smooth_loss}
\end{equation}
where $\dddot{\mathbf{p}}_{\tau}$ represents the jerk vector at the $\tau$-th sample point. 

The kinodynamic loss $\mathcal{L}_{\mathrm{kino}}$ effectively constrains the generated trajectory to remain within physical limits by bounding:
\begin{equation}
    \mathcal{L}_{\mathrm{kino}} = \frac{1}{N_\tau} \sum_{\tau=1}^{N_\tau} 
    \left( \big[ \|\mathbf{v}_{k,\tau}\|_2 - v_{\max} \big]_+^2 + \big[ \|\mathbf{a}_{k,\tau}\|_2 - a_{\max} \big]_+^2 \right),
\label{eq:kino_loss}
\end{equation}
\noindent where $\mathbf{v}_{k,\tau}$ and $\mathbf{a}_{k,\tau}$ denote the velocity and acceleration vectors, with $\|\cdot\|_2$ representing the $L_2$ norm. The operator $[\cdot]_+ = \max(\cdot, 0)$ ensures that penalties are only activated when the dynamical states violate the predefined physical limits $v_{\max}$ and $a_{\max}$.

Leveraging the calibrated TPGF via highly efficient GPU-accelerated batched operations, the overall learning procedure of the F2P framework is summarized in Algorithm~\ref{alg:f2p_training}.
\begin{table}
\centering
\caption{QUANTITATIVE BENCHMARKS IN VARIOUS SCENARIOS}
\label{tab:quantitative_comparison}

\renewcommand{\arraystretch}{1.4} 
\setlength{\tabcolsep}{1.5pt}     
\footnotesize                     

\begin{tabular}{llcccccc}
\toprule
\addlinespace[0.5ex]

\makecell[l]{\textbf{Configuration}} & 
\makecell[l]{\textbf{Method}} & 
\makecell{\textbf{SR} $\uparrow$} & 
\makecell{\textbf{TC} $\downarrow$} & 
\makecell{\textbf{SPL} $\uparrow$} & 
\makecell{\textbf{SNT} $\uparrow$} & 
\makecell{\textbf{ET} $\downarrow$} & 
\makecell{\textbf{RT} $\downarrow$} \\

\addlinespace[0.5ex]
\midrule

\multirow{5}{*}{\makecell[l]{\textbf{Scenario 1} \\ 25 Dynamic \\ 75 Static \\ Speed: 0.5 m/s}} 
& Ours               & \cellcolor{bestcolor}\textbf{1.00} & \cellcolor{bestcolor}\textbf{0.00} & \cellcolor{bestcolor}\textbf{0.96} & \cellcolor{bestcolor}\textbf{0.75} & 29.75 & \cellcolor{secondcolor}4.77 \\
& VIGO\cite{xu2022vision}         & \cellcolor{secondcolor}0.96 & \cellcolor{secondcolor}1.00 & \cellcolor{secondcolor}0.93 & 0.44 & 58.12 & 45.58 \\
& IPD-MPC\cite{xu2025intent}      & 0.76 & 7.00 & 0.68 & 0.72 & \cellcolor{bestcolor}\textbf{27.75} & 25.20 \\
& YOPO\cite{lu2024you}         & 0.92 & 2.00 & 0.91 & \cellcolor{secondcolor}0.73 & \cellcolor{secondcolor}28.81 & \cellcolor{bestcolor}\textbf{3.44} \\
& EGO-Planner\cite{zhou2020ego}   & 0.80 & 6.00 & 0.77 & 0.46 & 37.53 & 42.34 \\

\midrule

\multirow{5}{*}{\makecell[l]{\textbf{Scenario 2} \\ 75 Dynamic \\ 25 Static \\ Speed: 0.5 m/s}}
& Ours               & \cellcolor{bestcolor}\textbf{0.96} & \cellcolor{bestcolor}\textbf{1.00}  & \cellcolor{bestcolor}\textbf{0.92} & \cellcolor{bestcolor}\textbf{0.73} & \cellcolor{bestcolor}\textbf{29.00} & \cellcolor{secondcolor}4.65 \\
& VIGO\cite{xu2022vision}         & \cellcolor{secondcolor}0.80  & \cellcolor{secondcolor}6.00  & \cellcolor{secondcolor}0.78 & 0.30 & 59.62 & 49.67 \\
& IPD-MPC\cite{xu2025intent}      & 0.72  & 9.00  & 0.63 & 0.48 & 31.16 & 55.37 \\
& YOPO\cite{lu2024you}         & 0.68  & 10.00 & 0.68 & \cellcolor{secondcolor}0.62 & \cellcolor{secondcolor}30.18 & \cellcolor{bestcolor}\textbf{3.46} \\
& EGO-Planner\cite{zhou2020ego}   & 0.60  & 15.00 & 0.55 & 0.35 & 38.29 & 41.34 \\

\midrule

\multirow{5}{*}{\makecell[l]{\textbf{Scenario 3} \\ 75 Dynamic \\ 25 Static \\ Speed: 1.0 m/s}}
& Ours               & \cellcolor{bestcolor}\textbf{0.84} & \cellcolor{bestcolor}\textbf{4.00} & \cellcolor{bestcolor}\textbf{0.79} & \cellcolor{bestcolor}\textbf{0.61} & 31.12 & \cellcolor{secondcolor}4.52 \\
& VIGO\cite{xu2022vision}         & \cellcolor{secondcolor}0.68 & \cellcolor{secondcolor}9.00 & \cellcolor{secondcolor}0.62 & 0.31 & 58.37 & 44.40 \\
& IPD-MPC\cite{xu2025intent}      & 0.44 & 22.00 & 0.39 & 0.41 & \cellcolor{bestcolor}\textbf{26.62} & 32.99 \\
& YOPO\cite{lu2024you}         & 0.60 & 14.00 & 0.57 & \cellcolor{secondcolor}0.48 & \cellcolor{secondcolor}28.92 & \cellcolor{bestcolor}\textbf{3.45} \\
& EGO-Planner\cite{zhou2020ego}   & 0.28 & 35.00 & 0.26 & 0.17 & 37.20 & 45.05 \\

\bottomrule
\end{tabular}
\vspace{-12pt}
\end{table}

\section{EXPERIMENT}
In this section, we implement a series of experiments to evaluate the proposed method. All simulations are conducted on a workstation equipped with an Intel Core i9-14900KF CPU and an NVIDIA GeForce RTX 4090 GPU. As shown in Fig. \ref{fig:scenarios}, the data collection is executed within NVIDIA Isaac Sim\cite{xu2025navrl}, while subsequent benchmark comparisons are conducted in the ROS-Gazebo environment. The control architecture employs a hierarchical structure, coupling a cascaded PID outer loop with the PX4 inner loop\cite{chen2023flying}.

\subsection{Computational Efficiency Comparison}
We compare the computational efficiency of TPGF and ESDF under different batch sizes $B$ and trajectory sampling counts $N_\tau$. Using 2,000 randomly sampled offline frames, runtime is measured over 20 iterations after 5 warm-up runs. TPGF is computed in parallel on the GPU, whereas ESDF construction is performed on the CPU. As shown in Table \ref{tab:computation_efficiency}, TPGF is consistently more efficient than ESDF, particularly under batched evaluation, making it well suited for large-scale policy training in dynamic environments.
\begin{figure}
    \centering
    \includegraphics[width=1.0\linewidth, keepaspectratio]{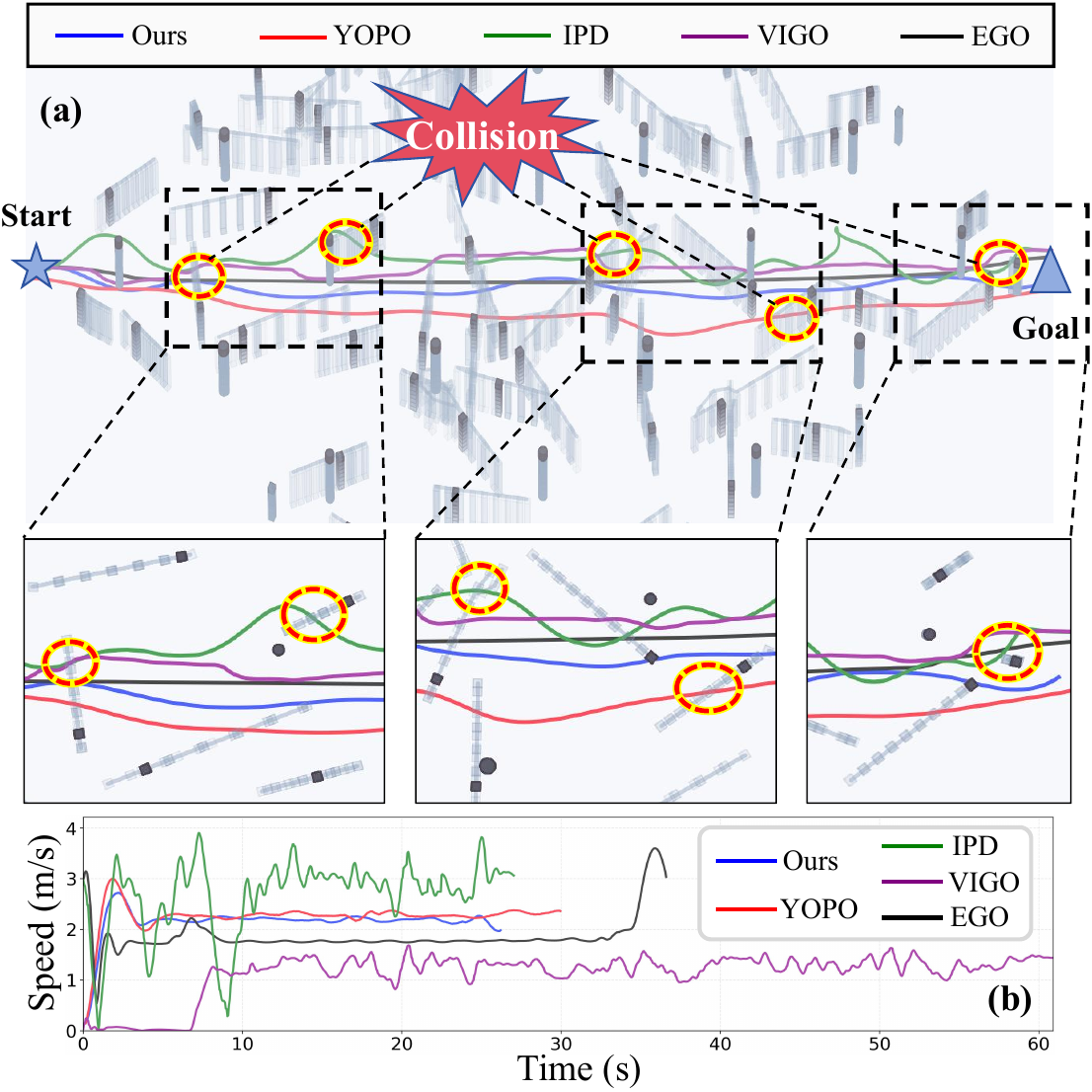}
    \vspace{-15pt}
    \caption{Visualization of a single flight trial for the evaluated methods. The methods share identical start and goal points, as well as consistent dynamic obstacle trajectories. (a) Spatiotemporal AAV trajectories, with enlarged views
    of three collision regions. The historical trajectories of moving obstacles are
    represented by connected transparent cuboids. (b) Velocity-time curves.}
    \label{fig:Benchmarks_flight}
    \vspace{-12pt}
\end{figure}

\subsection{Planning Benchmark Comparisons}
We employ widely used planning metrics for comparative evaluation\cite{xie2023drl, jiang2024bevnav}: 
\begin{itemize}
    \item \textit{Success Rate (SR)}, defined as the percentage of evaluation runs where the AAV safely reaches the target without any collisions (\%).
    \item \textit{Total Collision (TC)}, defined as the total number of collision events across all evaluation runs.
    \item \textit{Average Success Path Length (SPL)}, an integrated metric evaluating both success rate and trajectory efficiency.
    \item \textit{Average Success Navigation Time (SNT)}, a combined measure of time efficiency and success rate.
    \item \textit{Average Execution Time (ET)}, defined as the overall mean time spent executing each planning task (s).
    \item \textit{Average Replanning Time (RT)}, defined as the average computational latency required for each trajectory generation step (ms).
\end{itemize}

To comprehensively evaluate the performance of our proposed method, we benchmark it against four representative state-of-the-art (SOTA) vision-based planning approaches: 
\begin{table}
\centering
\caption{Ablation Study on Parameter Configurations and Planning Performance in Dynamic Scenarios}
\label{tab:ablation_combined}

\footnotesize
\renewcommand{\arraystretch}{1.2}
\setlength{\tabcolsep}{1.5pt}

\begin{tabular}{l c ccccc}
\toprule

\multirow{2}{*}{\textbf{Method}} &
\multirow{2}{*}{
    \makecell[c]{
        \textbf{Learned Parameters} \\
        $[\eta,\ \boldsymbol{s}_{c,b,h},\ \boldsymbol{w}_{c,b,h}]$
    }
} &
\multicolumn{5}{c}{\textbf{Planning Performance}} \\

\cmidrule(lr){3-7}

& &
\textbf{SR} $\uparrow$ &
\textbf{TC} $\downarrow$ &
\textbf{SPL} $\uparrow$ &
\textbf{SNT} $\uparrow$ &
\textbf{ET} $\downarrow$ \\

\midrule

Core only &
\makecell[c]{
    $[4.37,\{2.19,0.00,0.00\},$ \\
    $\{1.00,0.00,0.00\}]$
} &
0.20 & 48.00 & 0.18 & 0.16 &
\cellcolor{bestcolor}\textbf{28.32} \\

Body only &
\makecell[c]{
    $[4.40,\{0.00,9.26,0.00\},$ \\
    $\{0.00,1.00,0.00\}]$
} &
0.00 & 83.00 & 0.00 & 0.00 & 29.16 \\

Halo only &
\makecell[c]{
    $[4.39,\{0.00,0.00,10.42\},$ \\
    $\{0.00,0.00,1.00\}]$
} &
0.28 & 35.00 & 0.25 & 0.21 & 29.52 \\

\midrule

w/o $\mathcal{L}_{\mathrm{corr}}$ &
\makecell[c]{
    $[2.46,\{1.17,4.95,7.52\},$ \\
    $\{2.17,1.95,1.41\}]$
} &
0.04 & 66.00 & 0.03 & 0.03 &
\cellcolor{secondcolor}28.79 \\

w/o $\mathcal{L}_{\mathrm{safe}}$ &
\makecell[c]{
    $[2.46,\{1.18,4.96,7.51\},$ \\
    $\{2.16,1.95,1.41\}]$
} &
0.08 & 61.00 & 0.07 & 0.06 & 28.80 \\

w/o $\mathcal{L}_{\mathrm{grad}}$ &
\makecell[c]{
    $[4.15,\{2.29,9.55,10.66\},$ \\
    $\{0.60,3.85,4.97\}]$
} &
\cellcolor{secondcolor}0.76 &
\cellcolor{secondcolor}6.00 &
\cellcolor{secondcolor}0.71 &
\cellcolor{secondcolor}0.57 &
29.98 \\

\midrule

Ours &
\makecell[c]{
    $[4.28,\{2.21,9.38,10.53\},$ \\
    $\{0.68,3.96,5.06\}]$
} &
\cellcolor{bestcolor}\textbf{0.96} &
\cellcolor{bestcolor}\textbf{1.00} &
\cellcolor{bestcolor}\textbf{0.92} &
\cellcolor{bestcolor}\textbf{0.73} &
29.00 \\

\bottomrule
\end{tabular}

\vspace{2pt}

\parbox{\linewidth}{
\scriptsize
\textit{Note:}
$\boldsymbol{s}_{c,b,h}$ denotes the learned base Gaussian bandwidths
of the core, body, and halo layers in polar-grid cells, respectively.
The bucket-specific angular bandwidths are deterministically derived
from these base bandwidths through the range-dependent angular compensation
and are not independently learned.
}

\end{table}

\begin{figure}
    \centering
    \begin{minipage}{0.4939\linewidth}
        \centering
        \includegraphics[width=\linewidth, height=3.0cm]{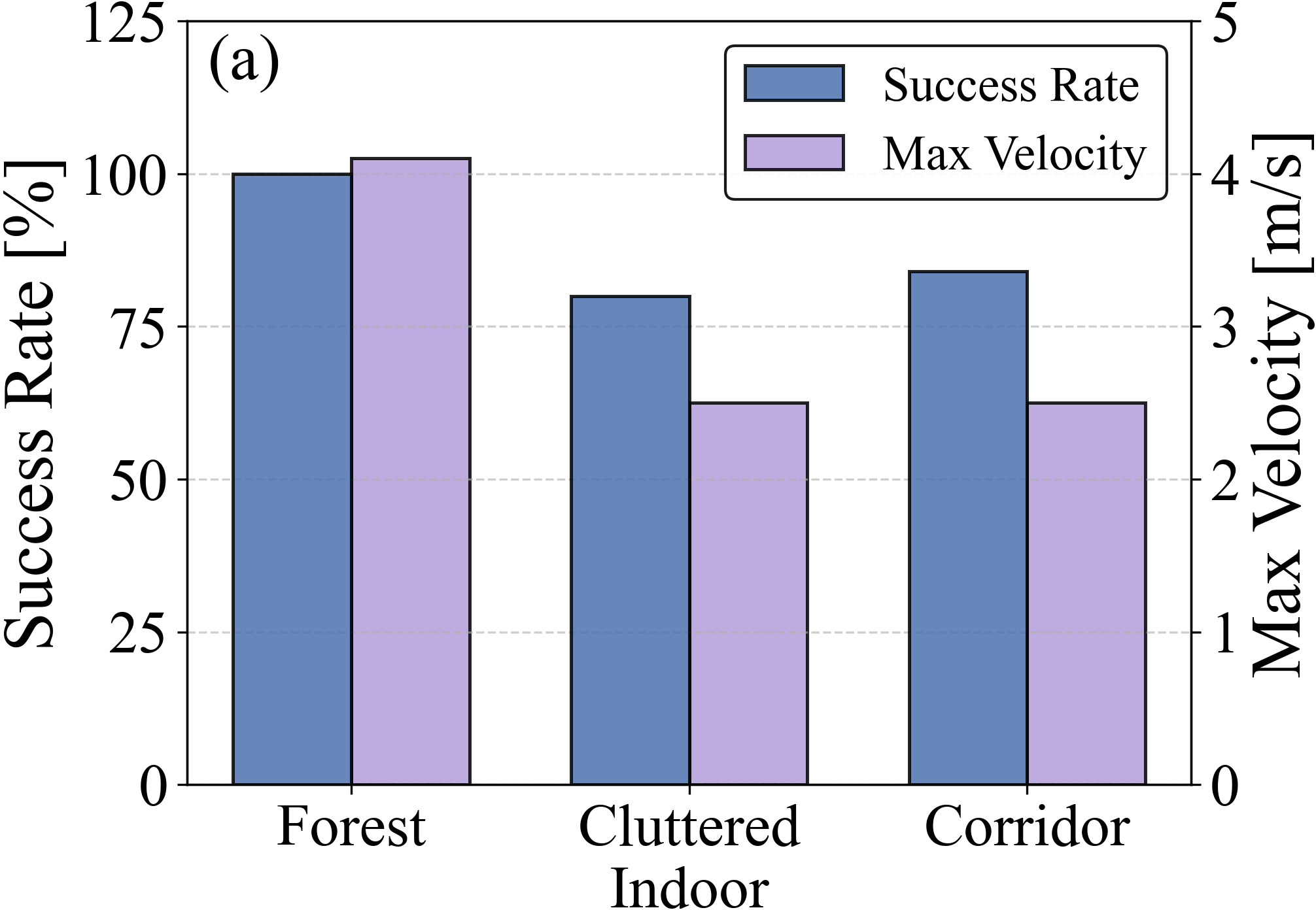}
        \vspace{0pt} 
        \centerline{\scriptsize (a) Generalization Results Statistics}
    \end{minipage}
    \hfill 
    \begin{minipage}{0.4939\linewidth}
        \centering
        \includegraphics[width=\linewidth, height=3.0cm]{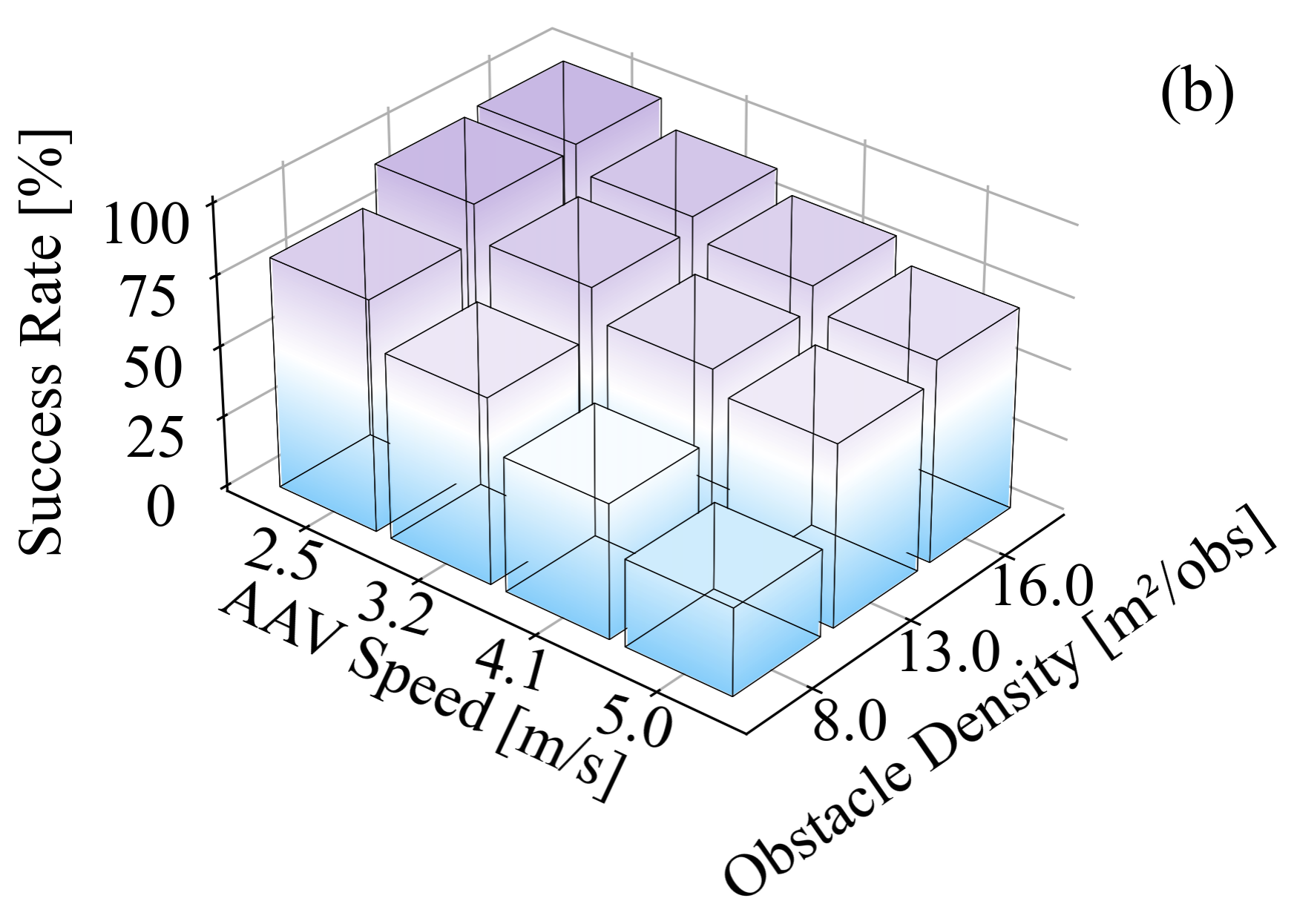}
        \vspace{0pt}
        \centerline{\scriptsize (b) Heat Map of Performance Boundaries}
    \end{minipage}
    
    \caption{Generalization and performance boundaries. (a) Success rates across diverse unseen scenarios. (b) Success rate degradation under scaled obstacle density and maximum velocity.}
    \label{fig:generalization}
    \vspace{-10pt}
\end{figure}
\begin{itemize}
    \item \textit{VIGO} \cite{xu2022vision}, a gradient-based B-spline trajectory optimization algorithm designed for dynamic scenes.
    \item \textit{IPD-MPC} \cite{xu2025intent}, an Intent Prediction-Driven MPC planner designed for dynamic scenes.
    \item \textit{YOPO} \cite{lu2024you}, an end-to-end planner tailored for static environments.
    \item \textit{EGO-Planner} \cite{zhou2020ego}, a classical modular planning method in static environments.
\end{itemize}

All methods are evaluated within a unified ROS-Gazebo simulation. As shown in Fig. \ref{fig:scenarios} (b), we construct a 40 $\times$ 40 m cluttered arena and design three distinct obstacle configurations to comprehensively assess robustness. For each method, the maximum flight velocity $v_{\max}$ is uniformly set to 2.5 m/s, and the evaluation metrics are calculated over 25 independent flight trials per scenario, with the scene randomly re-initialized at the beginning of every trial. 

Table \ref{tab:quantitative_comparison} demonstrates the consistent superiority of our method over the baselines in overall performance. Specifically, in both dense dynamic scenarios (average speeds of 0.5 m/s and 1.0 m/s), our approach achieves an exact 16.0\% improvement in success rate compared to the best-performing SOTA method (VIGO). Furthermore, it simultaneously maintains highly efficient average replanning times of under 4.80 ms across all test cases. These results confirm our method's superior effectiveness and agility in dynamic environments.
\begin{figure}
    \centering
    \begin{minipage}{0.4939\linewidth}
        \centering
        \includegraphics[width=\linewidth, height=2.7cm]{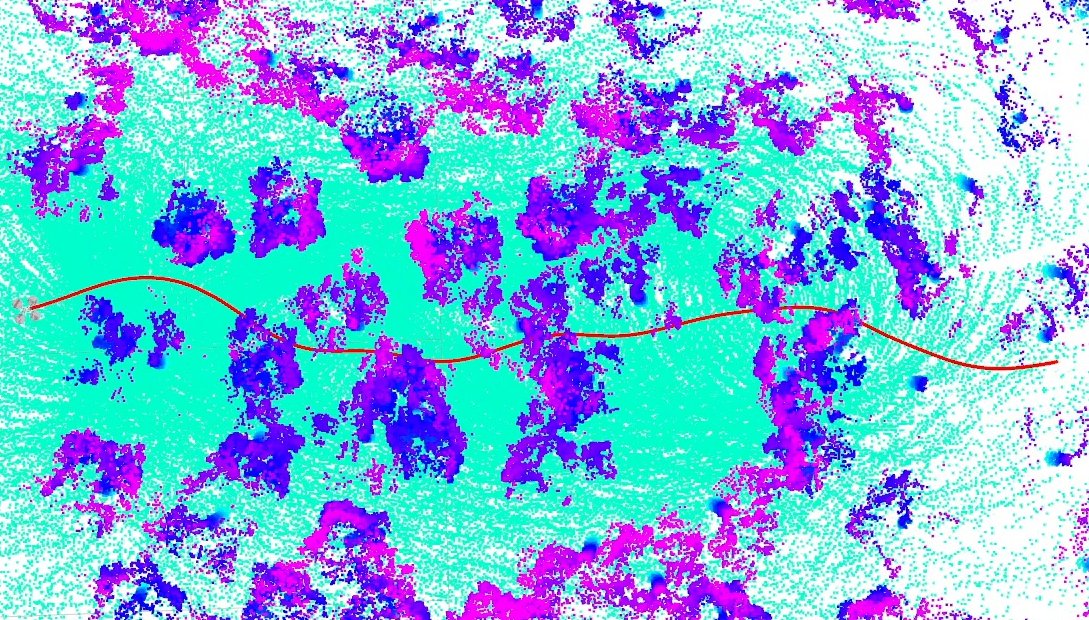}
        \vspace{3pt} 
        \centerline{\scriptsize (a) Trajectory in Static Forest}
    \end{minipage}
    \hfill 
    \begin{minipage}{0.4939\linewidth}
        \centering
        \includegraphics[width=\linewidth, height=2.7cm]{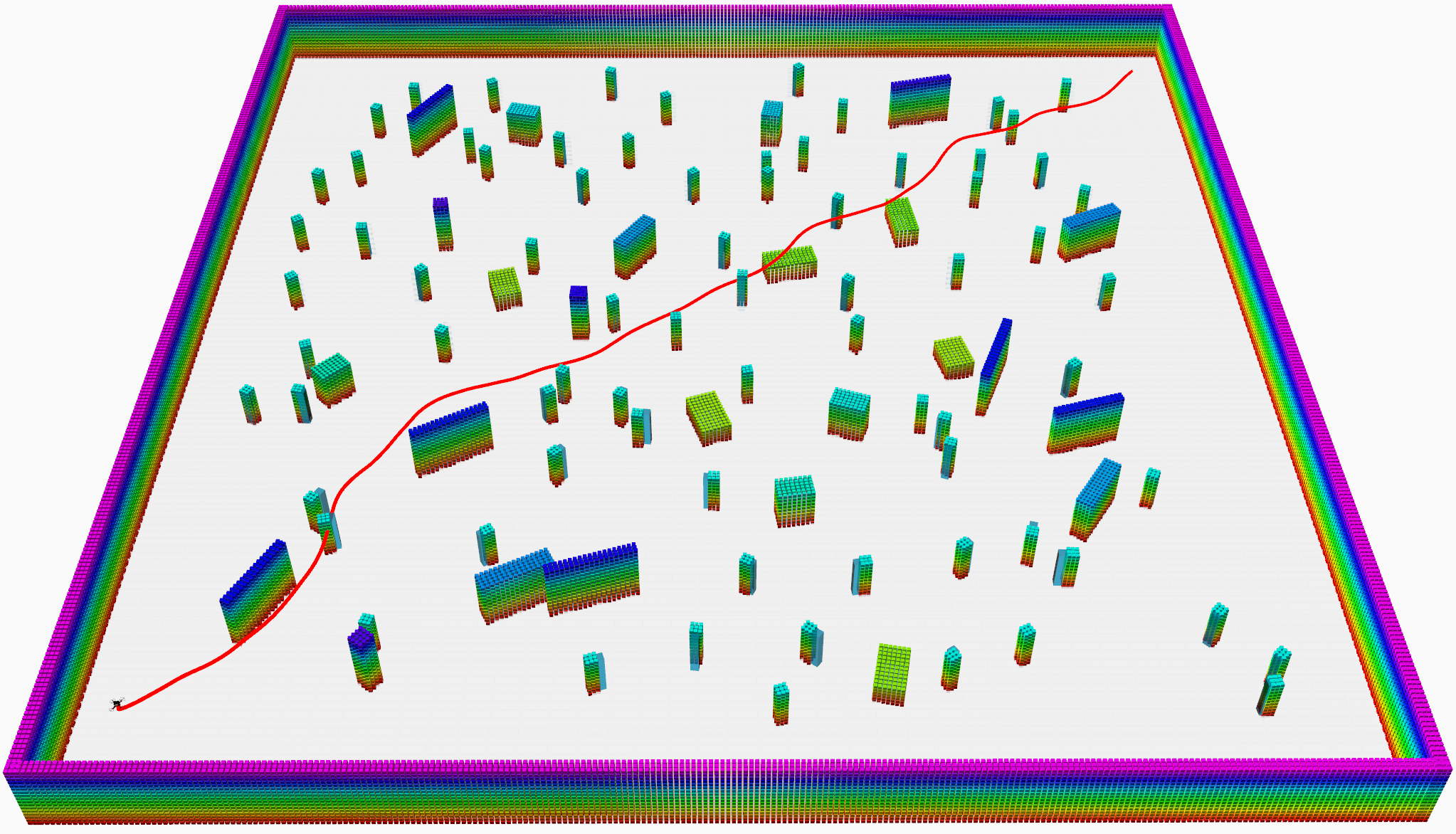}
        \vspace{3pt}
        \centerline{\scriptsize (b) Trajectory in Cluttered Indoor}
    \end{minipage}
    
    \vspace{3pt} 

    \begin{minipage}{\linewidth}
        \centering
        \includegraphics[width=1.0\linewidth, keepaspectratio]{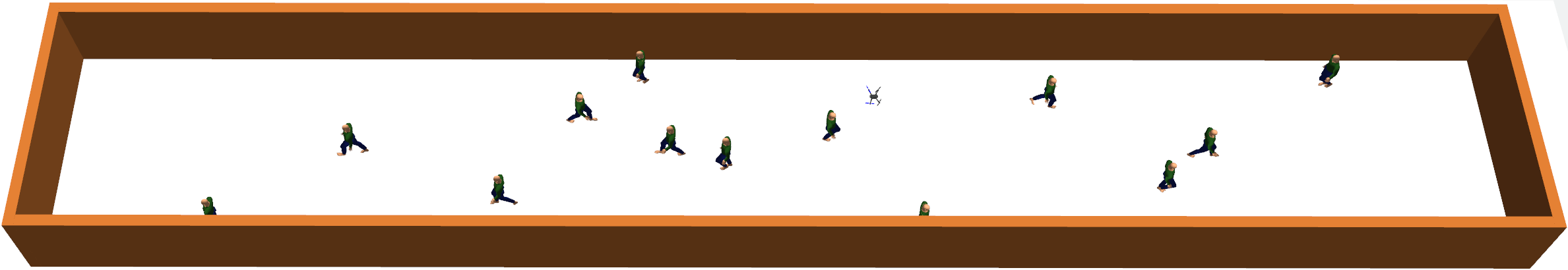} 
        
        \vspace{3pt} 
        
        \includegraphics[width=1.0\linewidth, keepaspectratio]{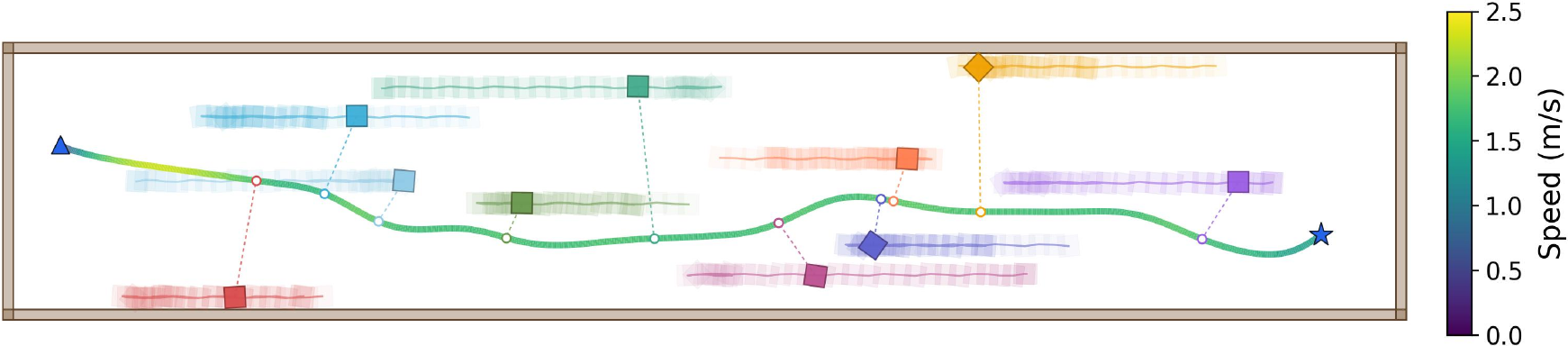} 
        
        \vspace{0pt}
        \centerline{\scriptsize (c) Gazebo and trajectories visualization in the crowded corridor}
    \end{minipage}
    
    \caption{AAV trajectories and generalization testing. The AAV successfully avoids all obstacles in the three cases. (a) and (b) illustrate the corresponding trajectories in the static forest and cluttered indoor. (c) demonstrates the Gazebo simulation and the corresponding AAV and dynamic-pedestrian trajectories. The AAV trajectory is color-coded by speed, while solid obstacle positions mark the closest approaches to the AAV during the flight.}
    \label{fig:trajectory_and_simulation}
    \vspace{-8pt}
\end{figure}
Fig. \ref{fig:Benchmarks_flight} illustrates a single trajectory sample alongside the corresponding velocity-time profiles for the five methods, evaluated under identical start and goal points and consistent dynamic obstacle trajectories. As highlighted by the red boxes in the figure, all baseline methods result in collisions, with the exception of Ours. Consistent with our analysis, the modular pipelines VIGO and IPD-MPC, which utilize DATMO to track dynamic obstacles, tend to exhibit conservative trajectory behaviors and lack smoothness. While YOPO and EGO-Planner generate smoother trajectories, their reliance on static-obstacle avoidance limits their ability to effectively evade active dynamic threats.

\subsection{Ablation Studies}
We conduct a series of ablation studies to systematically analyze the contribution of the multi-kernel Gaussian convolution and the carefully designed loss within the ESDF-guided offline learning framework. 
Specifically, to validate the structural necessity of the multi-scale formulation, we independently isolate the three kernels in Eq. \eqref{eq:PGF} to establish three single-kernel variants, where each variant is independently fine-tuned from the optimal configuration under the ESDF-guided framework. Furthermore, we independently omit the correlation loss $\mathcal{L}_{\mathrm{corr}}$ in Eq. \eqref{eq:loss_corr}, safety loss in $\mathcal{L}_{\mathrm{safe}}$ Eq. \eqref{eq:loss_safe}, and gradient loss in $\mathcal{L}_{\mathrm{grad}}$ Eq. \eqref{eq:loss_grad} from the optimization pipeline to learn the Gaussian kernel parameters.

Table \ref{tab:ablation_combined} summarizes the parameters and performance of these variants in a scenario with $25$ static and $75$ dynamic obstacles ($0.5 \text{ m/s}$).
The results indicate that single-kernel variants are inadequate for dynamic obstacle avoidance tasks. Regarding the loss constraints, the results demonstrate that $\mathcal{L}_{\mathrm{corr}}$ and $\mathcal{L}_{\mathrm{safe}}$ are indispensable for shaping a reliable multi-scale risk field and prove to be decisive for maintaining the baseline safety. Removing either loss degrades the TPGF into a narrow, core-dominated configuration with severely reduced body and halo weights, and the success rate drops near zero when either is omitted. Conversely, removing $\mathcal{L}_{\mathrm{grad}}$ results in minor parameter changes, while $\mathcal{L}_{\mathrm{grad}}$ acts as a key driver to further enhance trajectory quality and safety boundaries. This indicates that value-level constraints primarily shape the field's spatial distribution, whereas the gradient loss functions to enhance continuous directional guidance for policy optimization.
\begin{figure}
    \centering
    \includegraphics[width=1.0\linewidth, keepaspectratio]{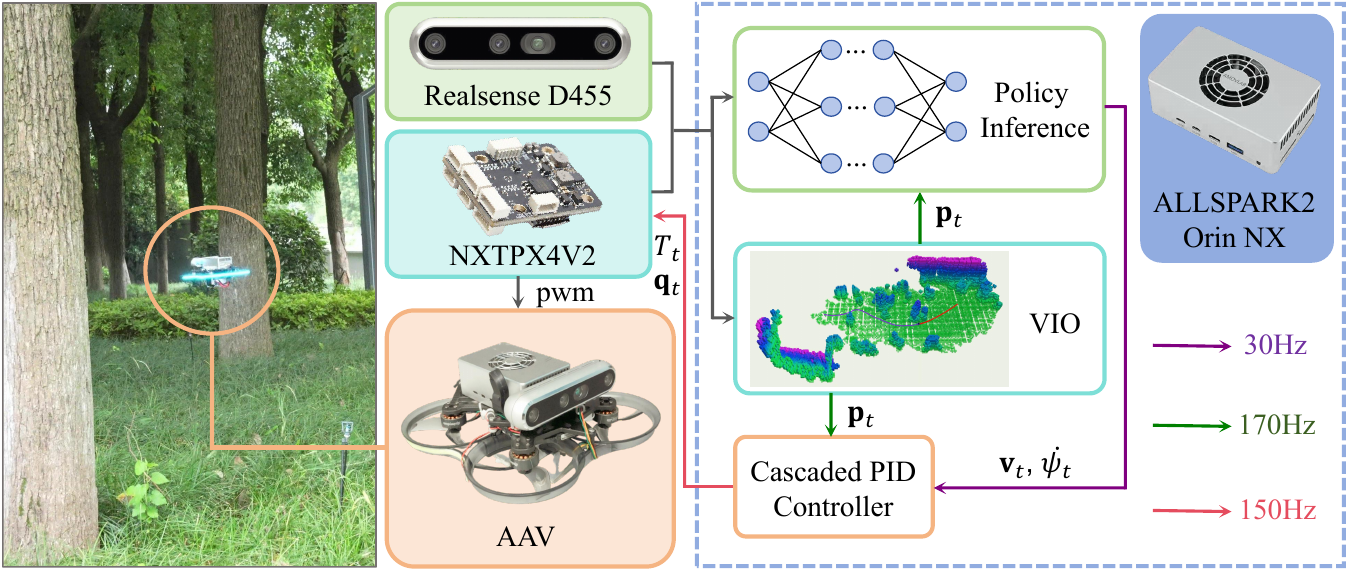}
    \vspace{-14pt}
    \caption{The real-world experimental system.}
    \label{fig:hardware}
    \vspace{-10pt}
\end{figure}

\subsection{Generalization Results}
To validate the generalizability of our framework, we conduct 50 independent flight trials across three distinct scenarios:
\begin{itemize}
    \item \textit{Static Forest}, a static forest scenario adapted from~\cite{lu2024you}, where Gaussian-distributed noise points are randomly scattered around the original cylindrical obstacles. The maximum AAV speed is set to 4.1\,m/s.
    \item \textit{Cluttered Indoor}, an indoor mixed scenario, where the flight space is enclosed by walls and populated with both large static obstacles and multiple moving cubic obstacles. The maximum AAV speed is set to 2.5\,m/s.
    \item \textit{Crowded Corridor}, an indoor corridor measuring 30\,m in length and 6\,m in width, populated with 13 dynamic pedestrian obstacles moving at an average speed of 0.6\,m/s. The maximum AAV speed is set to 2.5\,m/s.
\end{itemize}

The generalization results are summarized in Fig.~\ref{fig:generalization} (a). Without requiring any scenario-specific training or parameter fine-tuning, our method achieves a 100\% success rate in the unstructured \textit{Static Forest}, while maintaining robust performance when transitioning to constrained and dynamic environments. The corresponding flight trajectories across these distinct scenarios are visualized in Fig.~\ref{fig:trajectory_and_simulation}.

Fig.~\ref{fig:generalization} (b) illustrates the performance boundaries of the proposed method. To investigate these boundaries, we conduct comprehensive flight tests by jointly scaling the obstacle density and the maximum velocity of the AAV. During this phase, 200 independent trials are executed, with dynamic obstacles moving at an average speed of 0.5~m/s. 
The success rate exhibits an expected degradation trend as the environment becomes more cluttered and the flight speed increases. 

\subsection{Real World Experiments}
To demonstrate the sim-to-real transfer capability and robustness of the proposed method under real-world perceptual and dynamic uncertainties, we conduct a series of flight experiments in both a cluttered indoor environment and an unstructured dynamic forest scenario.
\begin{figure*}[htbp]
    \centering
    \vspace{10pt} 
    
    \definecolor{idxcolor_green}{RGB}{105, 155, 55}
    
    \definecolor{idxcolor_blue}{RGB}{70, 130, 180} 
    
    \newcommand{\idxblue}[1]{%
        \tikz \node[fill=idxcolor_blue, fill opacity=0.6, text opacity=1, text=white, inner sep=1.5pt] {\textbf{\scriptsize #1}};%
    }
    
    \newcommand{\idxgreen}[1]{%
        \tikz \node[fill=idxcolor_green, fill opacity=0.6, text opacity=1, text=white, inner sep=1.5pt] {\textbf{\scriptsize #1}};%
    }

    \begin{overpic}[width=1.0\linewidth]{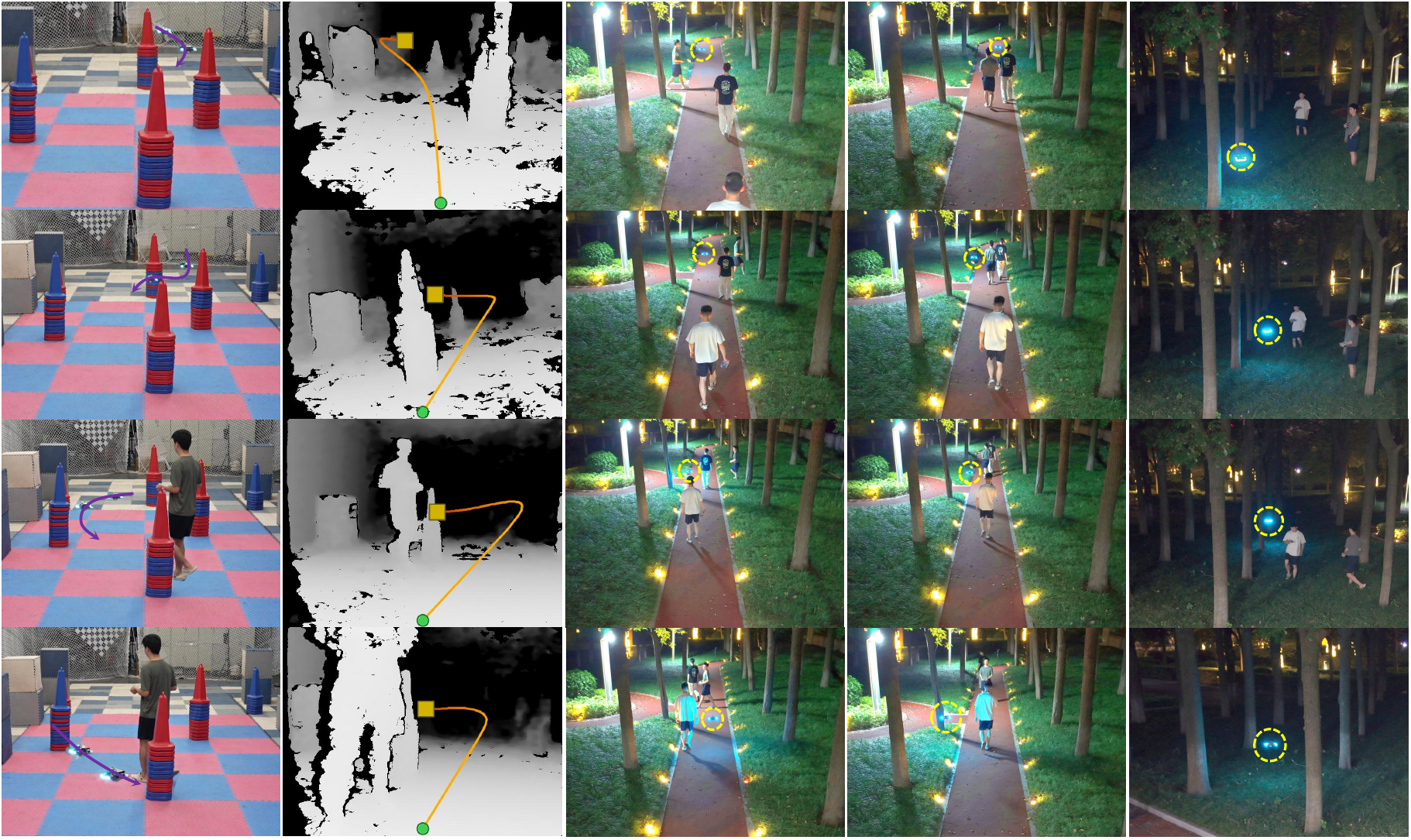}
        
        \put(16.5, 57.5){\idxblue{(a-1)}}
        \put(36.5, 57.5){\idxblue{(b-1)}}
        \put(56.5, 57.5){\idxgreen{(c-1)}}
        \put(76.5, 57.5){\idxgreen{(d-1)}}
        \put(96.5, 57.5){\idxgreen{(e-1)}}

        \put(16.5, 42.6){\idxblue{(a-2)}}
        \put(36.5, 42.6){\idxblue{(b-2)}}
        \put(56.5, 42.6){\idxgreen{(c-2)}}
        \put(76.5, 42.6){\idxgreen{(d-2)}}
        \put(96.5, 42.6){\idxgreen{(e-2)}}

        \put(16.5, 27.8){\idxblue{(a-3)}}
        \put(36.5, 27.8){\idxblue{(b-3)}}
        \put(56.5, 27.8){\idxgreen{(c-3)}}
        \put(76.5, 27.8){\idxgreen{(d-3)}}
        \put(96.5, 27.8){\idxgreen{(e-3)}}

        \put(16.5, 13.0){\idxblue{(a-4)}}
        \put(36.5, 13.0){\idxblue{(b-4)}}
        \put(56.5, 13.0){\idxgreen{(c-4)}}
        \put(76.5, 13.0){\idxgreen{(d-4)}}
        \put(96.5, 13.0){\idxgreen{(e-4)}}

    \end{overpic}

    \vspace{-5pt}
    \caption{A sequence of snapshots captures the quadrotor as it flies through different cluttered environments. (a) demonstrates evasive maneuvers against an actively approaching pedestrian in a cluttered indoor scene. (b) displays the corresponding depth observations and the real-time planned trajectories of the quadrotor. (c)-(e) showcase evasive flights in an outdoor forest scenario (with the quadrotor highlighted by a yellow circle), where (c) illustrates the continuous avoidance of pedestrians crossing both laterally and longitudinally; (d) depicts the quadrotor evading two pedestrians walking side-by-side while flying through the gap between a pedestrian and the trees; and (e) presents the continuous avoidance of multiple trees and pedestrians within the forest.}
    \label{fig:massive_matrix}
    \vspace{-7pt}
\end{figure*}
\begin{figure}[htbp]
    \centering
    \includegraphics[width=0.92\linewidth, keepaspectratio]{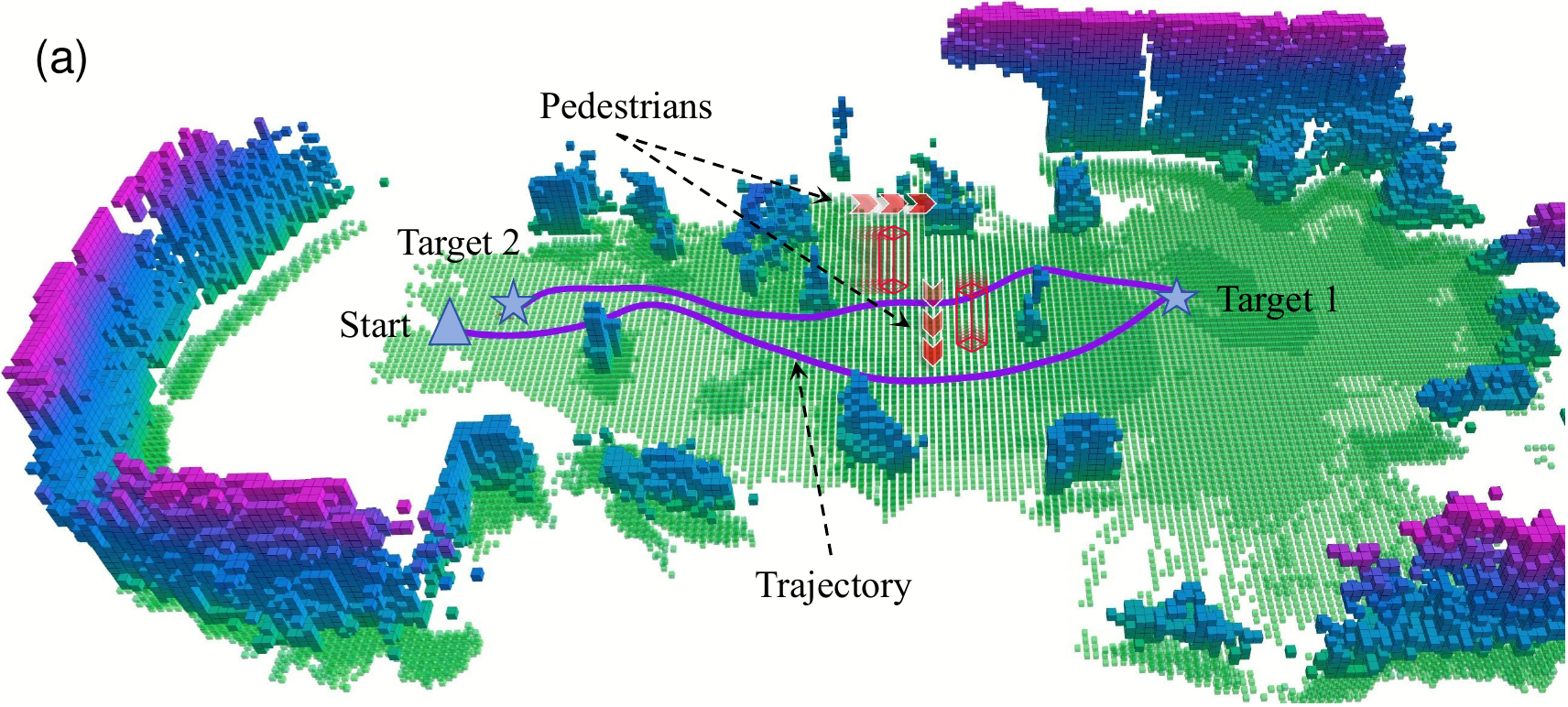}
    
    \vspace{4pt} 
    
    \includegraphics[width=0.88\linewidth, keepaspectratio]{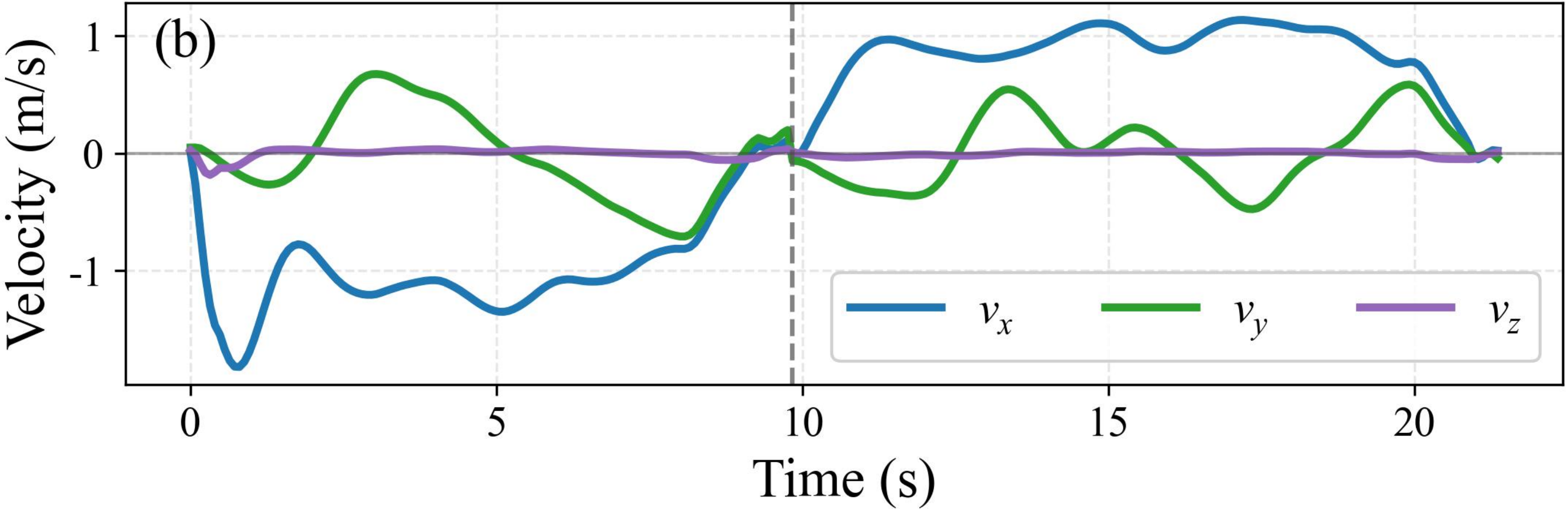}
    
    \vspace{-8pt} 
    \caption{Flight trajectory and the corresponding velocity profiles for the indoor scenario. (a) Executed 2D trajectory, where the purple curve represents the actual flight path of the quadrotor, blue voxels indicate static obstacles, and red cuboids denote dynamic pedestrians. (b) Corresponding velocity profiles over time. The 3-axis velocities are directly output from the odometry, representing the quadrotor's velocities in the world coordinate frame.}
    \label{fig:indoor_rviz}
    \vspace{-15pt}
\end{figure}
\subsubsection{Experimental Setup}
Fig. \ref{fig:hardware} illustrates the real-world experimental system, which comprises a custom-built quadrotor weighing 850~g with a 250~mm wheelbase, alongside its key onboard components. We employ an Intel RealSense D455 camera and the VINS framework\cite{qin2018vins} to provide high-frequency pose updates at 150~Hz. All high-level algorithms, including state estimation, cascaded controller and the proposed method, are run on an NVIDIA Jetson Orin NX. Our policy generates spatio-temporal polynomial trajectories, which are then translated into desired thrust and attitude commands via a cascaded PID controller\cite{chen2023flying} and subsequently sent to the NXTPX4V flight controller.
\begin{figure}
    \centering
    \includegraphics[width=0.98\linewidth, keepaspectratio]{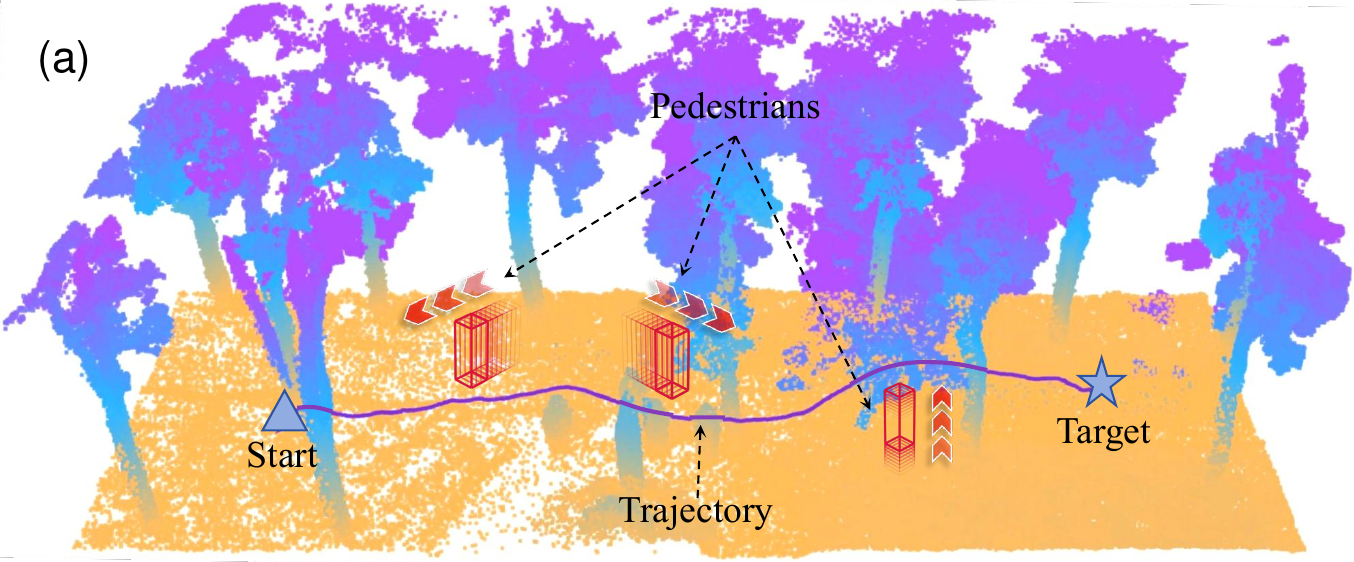}
    
    \vspace{6pt} 
    
    \includegraphics[width=0.88\linewidth, keepaspectratio]{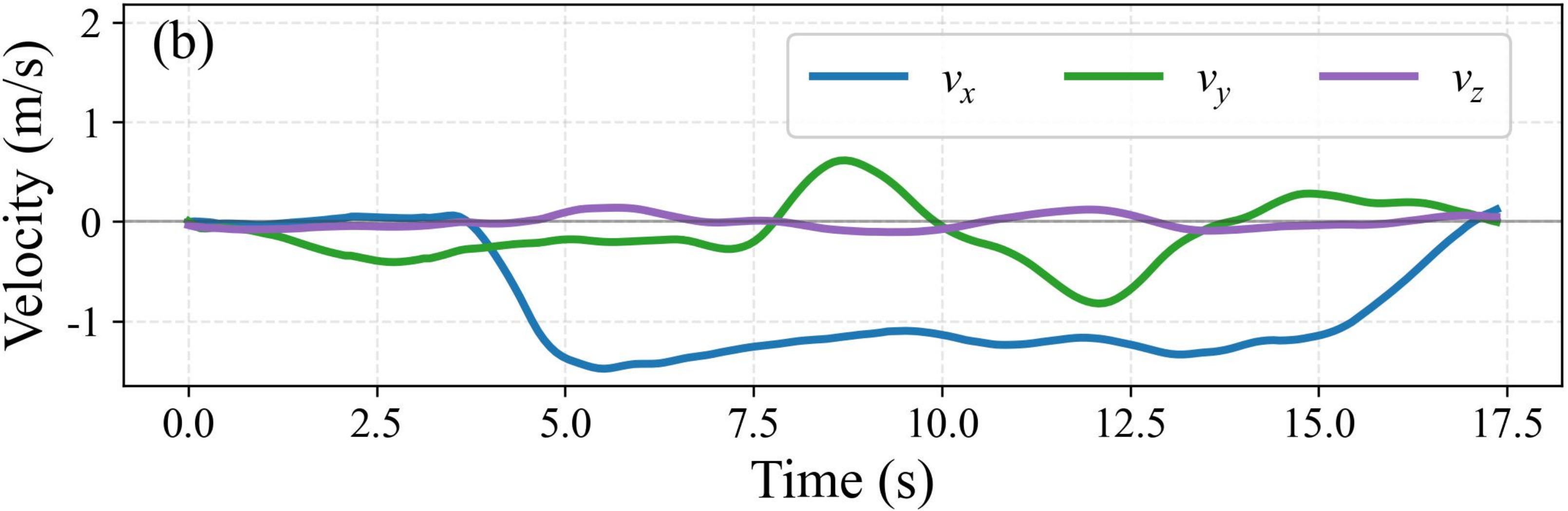}
    
    \vspace{-8pt} 
    \caption{Flight trajectory and the corresponding velocity profiles for the outdoor forest scenario.}
    \label{fig:outdoor_rviz}
    \vspace{-15pt}
\end{figure}
\subsubsection{Experimental Results}
To accelerate real-time onboard deployment, the policy network is optimized via TensorRT, achieving an inference latency of approximately 9 ms. Fig.~\ref{fig:massive_matrix} presents a sequence of snapshots that capture the quadrotor successfully avoiding obstacles in different scenarios. For pedestrian safety considerations, the maximum flight velocity is set to 1.6~m/s. Fig.~\ref{fig:indoor_rviz} and Fig.~\ref{fig:outdoor_rviz} respectively visualize the executed trajectories and the corresponding velocity profiles over time for the indoor and forest scenarios. These results confirm that the quadrotor generates smooth, collision-free trajectories, successfully flying through the obstacles.
In conclusion, these results demonstrate that the proposed method can achieve safe flights in cluttered and dynamic environments.

\section{CONCLUSION AND FUTURE WORK}
This paper proposes a two-stage F2P learning framework to provide accurate and efficient safety gradients for trajectory planning in dynamic environments. We introduce an EPO representation, and formulate TPGF via multi-kernel separable Gaussian convolution. Furthermore, we propose an ESDF-guided offline learning strategy to learn explicit geometric priors that effectively resolve inherent spatial ambiguity. This framework provides a unified representation of static and dynamic scenes, while supporting planning policy training in unstructured environments. Both simulations and real-world experiments demonstrate the highly competitive performance of the proposed framework.
Our future work will focus on embedding more robust and effective dynamic obstacle perception within our end-to-end framework. Additionally, we aim to enable real-time adaptive velocity adjustments by incorporating both obstacle and AAV velocities into the current purely spatial field, and ultimately extend the TPGF to three dimensions to enable robust 3D collision avoidance.
\bibliographystyle{IEEEtran} 
\bibliography{references} 

\end{document}